\documentclass[11pt]{article}

\usepackage[truedimen,margin=25truemm]{geometry}
\usepackage[numbers,compress]{natbib}

\usepackage[utf8]{inputenc} 
\usepackage[T1]{fontenc}    
\usepackage{hyperref}       
\usepackage{url}            
\usepackage{booktabs} 
\usepackage{amsfonts} 
\usepackage{nicefrac}
\usepackage{microtype} 
\usepackage{xcolor}

\usepackage{amsmath}
\usepackage{amssymb}
\usepackage{mathtools}
\usepackage{amsthm}

\usepackage{kasailab_def}
\usepackage{bm}

\usepackage{ascmac}
\usepackage{fancybox}

\usepackage{fancyhdr}

\usepackage{algorithm}
\usepackage{algorithmic}

\theoremstyle{plain}
\newtheorem{theorem}{Theorem}[section]

\theoremstyle{definition}
\newtheorem{definition}[theorem]{Definition}

\theoremstyle{remark}

\title{On the Convergence of Semi-Relaxed Sinkhorn with Marginal Constraint and OT Distance Gaps}

\author{Takumi Fukunaga \thanks{Department of Communications and Computer Engineering, School of Fundamental Science and Engineering, WASEDA University, 3-4-1 Okubo, Shinjuku-ku, Tokyo 169-8555, Japan (e-mail: f\_takumi1997@suou.waseda.jp) } \and Hiroyuki Kasai \thanks{Department of Communications and Computer Engineering, School of Fundamental Science and Engineering, WASEDA University, 3-4-1 Okubo, Shinjuku-ku, Tokyo 169-8555, Japan (e-mail: hiroyuki.kasai@waseda.jp)}}

\begin{document}

\maketitle

\begin{abstract}
This paper presents consideration of the Semi-Relaxed Sinkhorn (SR--Sinkhorn) algorithm for the semi-relaxed optimal transport (SROT) problem, which relaxes one marginal constraint of the standard OT problem. For evaluation of how the constraint relaxation affects the algorithm behavior and solution, it is vitally necessary to present the theoretical convergence analysis in terms not only of the functional value gap, but also of the {\it marginal constraint gap} as well as the {\it OT distance gap}. However, no existing work has addressed all analyses simultaneously. To this end, this paper presents a comprehensive convergence analysis for SR--Sinkhorn. After presenting the $\epsilon$-approximation of the functional value gap based on a new proof strategy and exploiting this proof strategy, we give the upper bound of the marginal constraint gap. We {also} provide its convergence to the $\epsilon$-approximation when two distributions are in the probability simplex. {Furthermore}, the convergence analysis of the OT distance gap to the $\epsilon$-approximation is given as assisted by the obtained marginal constraint gap. The latter two theoretical {results} are the first results presented in literature related to the SROT problem. 
\end{abstract}

\section{Introduction}
\label{sec:Intro}
The Optimal transport (OT) problem has attracted a surge of research interest because it expresses the distance between probability distributions, known as the Wasserstein distance \cite{Peyre_2019_OTBook,Villani_2008_OTBook}. This strong property enables us to apply this distance to widely diverse machine learning problems such as generative adversarial network \cite{arjovsky2017wasserstein}, graph optimal transport \cite{Togninalli_NIPS_2019, Huang_SigPro_2020}, clustering \cite{Kasai_ICASSP_2020, Fukunaga_ICPR_2020}, and domain adaptation \cite{Redko2019AISTATS,MLOT2021IJCAI}. These benefits nevertheless entail the shortcoming of high computational costs of solving the OT problem. Because the OT problem is formulated as convex linear programming problem \cite{Kantorovich_1942}, many dedicated solvers such as an interior-point method can yield solutions. However, its computational cost increases {\it cubically} in terms of the data size, which is prohibitive in large-scale applications. To alleviate high computational costs, the {\it entropy-regularized} OT problem has gained wide popularity because its differentiability and strong convexity of regularization terms enable construction of the celebrated {\it Sinkhorn algorithm} \cite{cuturiNIPS2013sinkhorn,sinkhornOriginal}. Theoretical analyses to obtain $\epsilon$-approximation solution have indicated that the Sinkhorn algorithm has the complexity of $\tilde{\mathcal{O}}(n^2/\epsilon^3)$ \cite{altschuler2017nearlinear}, which is subsequently improved with $\tilde{\mathcal{O}}(n^2/\epsilon^2)$ in \cite{Dvurechensky2018ICML}. Furthermore,  {\citet{Dvurechensky2018ICML}} {proposes} the greedy algorithm: {\it Greekhorn} algorithm. This algorithm achieves further improvements with $\tilde{\mathcal{O}}(n^2/\epsilon^2)$ \cite{LinAMDA2019ICML}. Moreover, the same authors propose {an} accelerated variant, and reveal the complexity $\tilde{\mathcal{O}}(n^{\frac{7}{3}}/\epsilon^{\frac{4}{3}})$ \cite{Lin2021efficiency}. Along another avenue of computational algorithms, several works point out that the tight {\it marginal} or {\it mass-conservation constraints} in the OT problem exacerbate degradation of the performance of some applications where weights need not be strictly preserved. To address these difficulties, the {\it relaxed} problem formulations have been proposed, which loosen the original marginal constraints. This category includes the partial optimal transport (POT) \cite{partialOT2010}, the fully relaxed optimal transport (FROT) or the unbalanced optimal transport (UOT) \cite{UOT2010Caff,chizat2017scaling}, and the semi-relaxed optimal transport (SROT) \cite{blondel2018smooth,NeurIPS2021robust}. Several reports have described that these approaches are more robust against outliers than the {standard} OT \cite{robust2020outlier,Killian2021minibatch}. Therefore, they have gained great success for applications such as generated adversarial networks \cite{robust2020outlier,outlierOT2022AISTAS,yang2019scalable}, domain adaptation \cite{Killian2021minibatch}, positive-unlabeled learning \cite{partial2020PU}, color transfer \cite{rabin2014addaptive}, and multi-label learning \cite{frogner2015learning}. However, they still exhibit a slow convergence property. To alleviate this slowness, faster algorithms have been developed using the Frank--Wolfe and block--coordinate approaches for the SROT problem \cite{Fukunaga_ICASSP_2022,fukunaga2021fast}. Recently, the Sinkhorn algorithm also has been extended to the UOT and the SROT {problems} \cite{NeurIPS2021robust,UOTSinkhorn2020}, which are called the Unbalanced Sinkhorn (UOT--Sinkhorn) and the Semi-Relaxed Sinkhorn (SR--Sinkhorn), respectively. 

This paper considers the SR--Sinkhorn algorithm for the KL-divergence penalized SROT problem with {an} entropy regularization. In general, to ascertain its algorithmic complexity, we discuss its convergence rate in terms of the functional value gap, or the solution gap. However, for {the} SROT problem, such analysis is insufficient to elucidate its comprehensive theoretical behaviors. 
One missing but necessary element is the convergence analysis of the {\it marginal constraint gap} because it provides knowledge of how far the relaxed marginal constraint deviates from the input probability when the algorithm terminates. It also enables us to {\it control} the degree of relaxation of the constraint, which is practically important from application perspectives of the SROT problem. 
Moreover, it is important to ascertain the deviation between the standard OT distance and that generated by an algorithm. Therefore, for the SROT problem with the probability simplex constraints, we consider projection of a solution of the SR--Sinkhorn algorithm onto the standard OT transport polytope, and denote the distance generated by such solution as the {\it SROT distance}. Under this setting, it is necessary to give the convergence analysis of the {\it OT distance gap} between the SROT distance and the OT distance. Although {\citet{NeurIPS2021robust}} has proved that the SR--Sinkhorn algorithm achieves $\tilde{\mathcal{O}}(n^2/\epsilon)$ with respect to the functional value gap, they do not discuss the marginal constraint gap and the OT distance gap. To provide better information about these topics, we provide a comprehensive theoretical result for the SR-Sinkhorn algorithm as follows. It should be emphasized that neither (ii) nor (iii) has been addressed in the literature related to the SROT problem. 
\\\\
\noindent
{\bf Our contribution{s}.} 

\begin{itemize}
\item[(i)]{\bf convergence analysis of {the} functional value gap based on new proof strategy:} We first provide a convergence analysis to the $\epsilon$-approximation in terms of the functional value gap based on a new proof strategy. The obtained complexity is the same order as that of \cite{NeurIPS2021robust} except constant numbers. However, our new proof strategy differentiates ours from those of \cite{NeurIPS2021robust,UOTSinkhorn2020} and other related papers in terms that we leverage the upper bound of {\it transport matrix} instead of {\it properties of objectives} as in \cite{NeurIPS2021robust, UOTSinkhorn2020}. This proof strategy is indeed straightforward and easy-to-follow. Beside all that, this strategy directly provides the succeeding two convergence results in (ii) and (iii). 

\item[(ii)]{\bf convergence analysis of the marginal constraint gap:} The theoretical deviation and the convergence of the relaxed marginal constraint gap are of great importance in the SROT problem. {\citet{extrapolation2022Quang}} presents the upper bound of the marginal constraint gap at optimal solutions under the KL-divergence penalized UOT problem with squared $l_2$-norm regularization. {\citet{Killian2021minibatch}} and {\citet{MinibatchPOT2020}} provide bounds on how the marginal constraint gap of the mini-batch UOT algorithm deviates from that of the standard UOT algorithm in a probabilistic way by the Hoeffding's inequality. However, they do not address the convergence rate to the $\epsilon$-approximation of the marginal constraint gap. By contrast, we provide the upper bound of the marginal constraint gap with the iteration number, and prove that it converges to a certain value. Then, assuming that two distributions $\vec{a}$ and $\vec{b}$ are the probability simplex, we obtain the convergence rate to the $\epsilon$-approximation in terms of the marginal constraint gap. This explicitly gives us the minimum iteration number of the SR--Sinkhorn algorithm such that the marginal constraint gap falls below $\epsilon$. It must be emphasized that {\citet{NeurIPS2021robust}, \citet{UOTSinkhorn2020}} and closely related papers explaining problems of SROT and UOT do not give {these} result{s}. 

\item[(iii)]{\bf convergence analysis of {the} OT distance gap:} {\citet{blondel2018smooth}} provides the upper bound of the optimal objective value gap separating the standard OT and the squared $l_2$-norm penalized UOT (and SROT). {\citet{UOTpenalized2021}} {describes} the possibility of approximation to the transport matrix of OT using regularization path algorithm for UOT. In fact, {\citet{extrapolation2022Quang}} {proves} the $\epsilon$-approximation to the OT distance gap using the extrapolation method for the KL-divergence penalized UOT with squared $l_2$-norm regularization. Nevertheless, no existing work has addressed the $\epsilon$-approximation of the OT distance gap in the SROT problem. To this end, {adding the probability simplex constraints on the probabilities, which are necessary for measuring a meaningful gap}, we provide the first $\epsilon$-approximation of the OT distance gap assisted by the marginal constraint gap obtained in (ii).

\end{itemize}

The paper is organized as follows. {\bf Section 2} presents preliminary descriptions of optimal transport, relaxed optimal transport, and the Sinkhorn algorithms. Furthermore, {\bf Section 3} presents the formulations of the SROT problems and the SR Sinkhorn algorithm. In {\bf Section 4}, our main theoretical results related to the SR--Sinkhorn algorithm are presented.  {\bf Section 5} shows numerical analysis. It is noteworthy that the Robust Semi Sinkhorn (RS--Sinkhorn) algorithm in \cite{NeurIPS2021robust} is identical to SR--Sinkhorn. Consequently, we use SR--Sinkhorn instead of the RS--Sinkhorn throughout this paper.

\section{Preliminaries}
\label{sec:Preliminaries}
$\mathbb{R}^n$ denotes $n$-dimensional Euclidean space, and $\mathbb{R}^n_+$ denotes the set of vectors in which all elements are non-negative. $\mathbb{R}^{m \times n}$ represents the set of $m \times n$ matrices. Also, $\mathbb{R}^{m \times n}_+$ stands for the set of $m \times n$ matrices in which all elements are non-negative. $\Delta^n$ represents the probability simplex as $\Delta^n = \lbrace \vec{x} \in \mathbb{R}^n : \vec{x}_i \geq 0 , \sum_i \vec{x}_i = 1 \rbrace$. We present vectors as bold lower-case letters $\vec{a},\vec{b},\vec{c},\dots$ and matrices as bold-face upper-case letters $\mat{A},\mat{B},\mat{C},\dots$. The $i$-th element of $\vec{a}$ and the element at the $(i,j)$ position of $\mat{A}$ are represented respectively as $\vec{a}_i$ and $\mat{A}_{i,j}$. In addition, $\vec{1}_n \in \mathbb{R}^n$ is the $n$-dimensional vector in which all the elements are one. $\vec{\delta}_{x}$ is the Delta function at position $x$. For \vec{x} and \vec{y} of the same size, $\langle \vec{x},\vec{y} \rangle = \vec{x}^T\vec{y}$ is the Euclidean dot-product between vectors. For two matrices of the same size \mat{A} and \mat{B}, $\langle \mat{A},\mat{B}\rangle={\rm tr}(\mat{A}^T\mat{B})$ is the Frobenius dot-product. For a vector $\vec{x}$, the $i$-th element of $\exp (\vec{x})$ and $\log (\vec{x})$ respectively represent $\exp (\vec{x}_i)$ and $\log (\vec{x}_i)$. $\mathrm{KL}(\vec{x},\vec{y})$ stands for the KL divergence between $\vec{x} \in \mathbb{R}_+^n$ and $\vec{y} \in \mathbb{R}_+^n$, which is defined as $\sum_i \vec{x}_i \log {(\vec{x}_i/\vec{y}_i)} - \vec{x}_i + \vec{y}_i$. $\mathrm{H}(\mat{T})$ represents the entropy term as $\mathrm{H}(\mat{T})=-\sum_{i,j}\mat{T}_{i,j}(\log \mat{T}_{ij}-1)$. $a = \mathcal{O}(f(n,\epsilon))$ expresses the {upper bound} satisfying the inequality $a \leq C \cdot f(n,\epsilon)$, where $C$ is independent of $n,\epsilon$. $a = \tilde{\mathcal{O}} (f(n,\epsilon))$ stands for the previous inequality for the constant $C$, which is dependent of $n,\epsilon$. Herein, $n$ and $\epsilon$ respectively represent a dimension and an approximation constant.

\subsection{Optimal transport}
The Kantorovich relaxation formulation of the optimal transport (OT) problem \cite{Kantorovich_1942} is explained briefly. Let $\vec{a}$ and $\vec{b}$ be {probabilities} or positive weight vectors as $\vec{a}=(\vec{a}_1, \vec{a}_2, \ldots, \vec{a}_m)^T \in \mathbb{R}_+^m$ and $\vec{b}=(\vec{b}_1, \vec{b}_2, \ldots, \vec{b}_n)^T \in \mathbb{R}_+^n$, respectively. Given two empirical distributions, i.e., {discrete measures}, $\vec{\nu}=\!\sum_{i=1}^m a_i\vec{\delta}_{{x}_i}$, $\vec{\mu}=\!\sum_{j=1}^n b_j\vec{\delta}_{{y}_j}$ and the ground cost matrix $\mat{C} \in \mathbb{R}^{m \times n}$ between their supports, the problem can be formulated as
\begin{equation}
	\label{eq:FormulationOptimalTransport}
	 \defmin_{\scriptsize {\mat{T} \in \mathcal{U}(\vec{a},\vec{b})}}\ \langle \mat{C} ,\mat{T}\rangle,
\end{equation}
where $\mat{T} \in \mathbb{R}^{m \times n}$ represents the {\it transport matrix}, and where the domain $\mathcal{U}(\vec{a},\vec{b})$ is defined as 
\begin{equation}
	\label{eq:TransportPolytope}
	\mathcal{U}(\vec{a},\vec{b}) = \Bigl\{ \mat{T} \in \mathbb{R}^{m \times n}_{+}: \mat{T}\vec{1}_n = \vec{a},\mat{T}^{T}\vec{1}_m = \vec{b} \Bigr\},
\end{equation}
where $\mat{T}\vec{1}_n = \vec{a}$ and $\mat{T}^{T}\vec{1}_m = \vec{b}$ are the {\it marginal constraints}. Moreover, we present the sum of the two vectors respectively as $\alpha > 0$ and $\beta > 0$, i.e., $\sum_{i=1}^m {\vec{a}_i} = \alpha$ and $\sum_{i=1}^n {\vec{b}_i} = \beta$. Note that $\alpha$ is equal to $\beta$ in the standard OT formulation. The obtained OT matrix $\mat{T}^*$ brings powerful distances as $\mathcal{W}_p(\vec{\nu},\vec{\mu}) = \langle \mat{T}^*,\mat{C} \rangle^{\frac{1}{p}}$, which is known as the $p$-th order {\it Wasserstein distance} \cite{Villani_2008_OTBook}. It is used in various fields according to the value of $p$. Especially, the distance is applied to computer vision \cite{Levina_ICCV_2001} when $p=1$, and to clustering \cite{cuturi14barycenter} when $p=2$. Throughout this paper, when $\mat{C}$ is the {ground cost} matrix and $p=1$, we specifically designate the $1$-th order Wasserstein distance as the {\it OT distance}.

\subsection{Relaxed optimal transport}
\label{Sec:ROT}
As described above, the OT problem is generally difficult to solve efficiently. In addition, the strict marginal constraints of the OT formulation might engender difficulties in many applications. To address these issues, the relaxed OT problems have attracted attention over the years. This subsection introduces variants of such relaxed OT problems in three ways.
\\\\
\noindent{\bf Partial Optimal Transport.} The partial optimal transport (POT) problem, which relaxes the domain of the constraints \cite{ferradans2013regularized}, is proposed to address the degradation of the strict constraints and to control the mass between two points. A noteworthy point is that the relaxed domain retains the linear constraints as the standard OT problem. For that reason, existing solvers of linear programming are applicable. {\citet{partial2020PU}} convert the POT problem into an augmented form by introducing the slack variable. The control of their variable enables the improvement of positive unlabeled learning problems. Another proposed problem is to remove one constraint as
\begin{eqnarray}
\label{Eq:Sinkhorn}
	\defmin_{{{\bf T}\geq \bm{0},{\bf T}^T\bm{1}_m = \bm{b}}}\ \langle \mat{C} ,\mat{T} \rangle.	
\end{eqnarray}
This solution is the summation of minimum costs of each row or column vector. Therefore, they are solvable faster than solving linear programming problem. This idea is applicable to, for example, style transfer problems \cite{kolkin2019style,FaststyleTransfer}. 
\\\\
\noindent{\bf Fully Relaxed or Unbalanced Optimal Transport.} As another line of attempt, the penalty of the domains defined in (\ref{eq:TransportPolytope}) is added to the objective function as regularizers \cite{blondel2018smooth}. Relaxation of the marginal constraints is effective when only partial transport is allowed. Relaxing both marginal constraints in (\ref{eq:TransportPolytope}) yields the following relaxed formulation as
\begin{equation*}
	\label{eq:SmoothRelaxedOptimalTransport}
	\defmin_{\scriptsize {\mat{T}\geq \vec{0}}}\ \langle \mat{C} ,\mat{T} \rangle + \frac{1}{2}\Phi(\mat{T}\vec{1}_n,\vec{a})+\frac{1}{2}\Phi(\mat{T}^T\vec{1}_m,\vec{b}),
\end{equation*}
where $\Phi(\vec{x}, \vec{y})$ is a smooth divergence measure function. This problem, designated as the {\it fully relaxed} OT (FROT) problem, or the {\it unbalanced} OT (UOT) problem, has attracted attention in various fields. In fact, {\citet{Killian2021minibatch}} have proposed a mini-batch-based algorithm, and performed a related theoretical analysis. The UOT problem on tree is also proposed \cite{UOTtree2020}. Recently, the UOT problem has been recast as a non-negative penalized linear regression problem  solved using a variant of majorization-minimization algorithms \cite{UOTpenalized2021}. In \cite{extrapolation2022Quang}, the UOT with $l_2$ squared regularization is proposed; it is solved using the extrapolation method.
\\\\
\noindent{\bf Semi-Relaxed or Semi-Constrainted Optimal Transport.} We also have an alternative formulation, which relaxes only one of the two marginal constraints in (\ref{eq:TransportPolytope}). This is designated as {\it semi-relaxed} OT (SROT) problem, or the {\it semi-constrained} OT (SCOT), which is formally defined as
\begin{equation}
	\label{eq:SmoothSemiRelaxedOptimalTransport}
	\defmin_{\scriptsize {\mat{T}\geq \vec{0},\mat{T}^T\vec{1}_m = \vec{b}}}\ \langle \mat{C} ,\mat{T} \rangle + \Phi(\mat{T}\vec{1}_n,\vec{a}).
\end{equation}
This setting is useful, for example, with color transfer problems \cite{blondel2018smooth}. {\citet{rabin2014addaptive}} also proposed the weighted regularization term $\|\kappa - \vec{1}_n\|_1$ and the relaxed weighted OT so that the ratio of the source image approaches that of the reference image. Recently, the SROT formulation has been applied for graph dictionary learning \cite{vincentcuaz2021semirelaxed}. It also exhibits the robustness to outliers in generative models \cite{robust2020outlier,yang2019scalable}. As a solver development side, a fast block--coordinate Frank--Wolfe (BCFW) algorithm is proposed for the SROT problem with $\Phi(\vec{x}, \vec{y})=\frac{1}{2}\| \vec{x}-\vec{y}\|^2_2$ \cite{Fukunaga_ICASSP_2022,fukunaga2021fast}, where the upper bounds of the worst convergence iterations are provided along with equivalence between the linearization duality gap and the Lagrangian duality gap.

\subsection{Entropy regularization and Sinkhorn algorithm}
\label{sec:Sinkhorn}
A different, but more popular line of algorithms without relaxation of the standard OT problem are entropy-regularized approaches. Among them, the most popular algorithm is the Sinkhorn algorithm \cite{sinkhornOriginal}, which is faster and which enables a parallel implementation because of the differentiability and the strong convexity of entropy term. In addition, the resultant modified OT distance is effective in various machine learning problem \cite{cuturiNIPS2013sinkhorn}. Furthermore, {\citet{chizat2017scaling}}  proposed stabler variants which cope with its numerical unsuitability and low robustness against small values of the regularizer, yet they have adversely slow convergence. 

Later, many researchers explored analysis of the Sinkhorn algorithm. {\citet{altschuler2017nearlinear}} show that the complexity is $\mathcal{O}(n^2/\epsilon^3)$ in terms of $\ell_1$ norm error of the OT constraints and its complexity is improved to be $\mathcal{O}(n^2/\epsilon^2)$ in \cite{Dvurechensky2018ICML}. Furthermore, the greedy algorithm, {\it Greekhorn} is developed in \cite{altschuler2017nearlinear}. Later, its complexity is proved to be $\mathcal{O}(n^2/\epsilon^2)$ in \cite{LinAMDA2019ICML}. The same authors propose the accelerated algorithm, which has $\mathcal{O}(n^{\frac{7}{3}}/\epsilon^{\frac{4}{3}})$ \cite{Lin2021efficiency}. Along another avenue of development, variants of the primal-dual method have been proposed {\cite{Dvurechensky2018ICML,LinAMDA2019ICML,Guo2020APDRCR}}. Several methods of total complexities $\mathcal{O}(n^2/\epsilon)$ are developed \cite{blanchet2020optimal,jambulapati2019direct}. It is true, however, that these implementations are difficult using the second-order and maximum flow. There are also an inexact proximal point method using the KL divergence \cite{xie2020sinkhorn} and an alternative minimization method introducing Nesterov's acceleration \cite{Melia2021alternate}.

This entropy regularization has also been applied to the UOT problem. The corresponding Sinkhorn-like algorithm {called the UOT--Sinkhorn algorithm has been} developed \cite{UOTSinkhorn2020}. {\citet{chizat2017scaling}} present a convergence rate with respect to the Thompson metric. Similar works provide analyses of the convergence rate and the total complexity $\tilde{\mathcal{O}}(n^2/\epsilon)$ in terms of the maximum norm \cite{janati20aSequence,UOTSinkhorn2020}. The multi-marginal POT has a complexity $\tilde{\mathcal{O}}(m^3(n+1)^m/\epsilon^2)$ \cite{MultimarginalOT2022AISTATS}. For a large relaxation parameter, the UOT--Sinkhorn is slow because the linear rate $(1+\frac{\epsilon}{\tau})^{-1}$ approaches $1$. For this issue, the accelerated Sinkhorn algorithm for UOT and the $1$-D case Frank--Wolfe for the UOT have been proposed \cite{UOT2022FrankWolfe}. {As for the SROT problem, the SR--Sinkhorn algorithm or the RS--Sinkhorn algorithm has been developed, and its complexity is proved to be ${\tilde{\mathcal{O}}}(n^2/\epsilon)$ in \cite{NeurIPS2021robust}}.

\section{SR--Sinkhorn algorithm}
\label{sec:SRSinkhorn}
Addressing the KL divergence for $\Phi(\cdot)$ in (\ref{eq:SmoothSemiRelaxedOptimalTransport}), we formally define a SROT problem with the KL divergence as
\begin{equation}
	\label{eq:KLSemiRelaxed}
	\min_{{\bf T}\geq \bm{0}, {\bf T}^T\bm{1}_n=\bm{b}}
	\Bigl\{ f(\mat{T}) :=\langle \mat{C} ,\mat{T} \rangle + \tau\mathrm{KL}(\mat{T}\vec{1}_n,\vec{a})\Bigr\},
\end{equation}
where $\tau >0$  is a regularization parameter. We then define an entropy regularized SROT problem as
\begin{equation}
	\label{eq:EntropySemiRelaxed}
	{\defmin_{{\bf T}\geq \bm{0}, {\bf T}^T\bm{1}_n=\bm{b}}}
	\Bigl\{ g(\mat{T}) :=\langle \mat{C} ,\mat{T}\rangle + \tau\mathrm{KL}(\mat{T}\vec{1}_n,\vec{a})-\eta \mathrm{H}(\mat{T})\Bigr\},
\end{equation}
where $\eta >0$  is a regularization parameter. We denote the optimal solution, i.e., the optimal transport matrix, of (\ref{eq:EntropySemiRelaxed}) as $\mat{T}^*$, i.e., $\mat{T}^*:=\defargmin g(\mat{T})$, and denote $\mat{T}^*\vec{1}_n$ and $(\mat{T}^*)^T\vec{1}_n$ as $\vec{a}^*$ and $\vec{b}^*$, respectively.
Under this formulation, considering the dual form of (\ref{eq:EntropySemiRelaxed}), the SR--Sinkhorn algorithm can be derived \cite{NeurIPS2021robust}. We define $\vec{u}$ and $\vec{v}$ as dual variables of the dual form of (\ref{eq:EntropySemiRelaxed}), and present $\vec{u}^*$ and $\vec{v}^*$ as the optimal solutions of the dual form. We also designate the updates of $\vec{u}$ and $\vec{v}$ in the algorithm, respectively, as the even update and the odd update. Detailed deviation of the algorithm is given in the supplementary material. The pseudocode of the SR--Sinkhorn algorithm is also shown in the supplementary material. 

\section{Main results}
\label{sec:Main}
This section presents our main theoretical results related to the SR--Sinkhorn algorithm. The first result in {\bf Section \ref{Sec:FuncAnalysis}} is the total complexity with respect to the functional value {gap}. The obtained results resemble {those} of \cite{NeurIPS2021robust}. It is nevertheless noteworthy that our emphasis lies on our {\it new proof strategy}, which successfully derives the new theoretical results in the succeeding subsections. {\bf Section \ref{Sec:MarginalAnalysis}} presents the theoretical upper bound and the convergence rate of the relaxed marginal constraint gap, which are of great importance in the relaxed OT problems. {\bf Section \ref{Sec:OTDistanceAnalysis}} evaluates the OT distance gap. The latter two analyses have not been addressed in the literature of the SROT problem. Full proofs of all results described below are presented in the supplementary material.

For our analysis, we formally define the {\it marginal constraint gap} and the {\it OT distance gap}:
\begin{definition}[marginal constraint gap and OT distance gap]
We denote the weight vector $\vec{a}$ and the transport matrix $\mat{T}$ at the $k$-th iteration by the SR--Sinkhorn algorithm as $\vec{a}^{(k)}$ and $\mat{T}^{(k)}$, respectively. The marginal constraint gap is defined as $\|\vec{a}^{(k)} - \vec{a}\|$. Representing the projected matrix of $\mat{T}^{(k)}$ using \cite[Algorithm 2]{altschuler2017nearlinear} as $\mat{Y}$, we denote $\langle \mat{C}, \mat{Y} \rangle$ as the {\it SROT distance} at the $k$-th iteration or simply the {\it SROT distance}. Then, the deviation between the SROT distance and the OT distance, i.e., $\langle  \mat{C} , {\mat{Y}} - \mat{T}^{\rm OT} \rangle$, is {defined as} the {\it OT distance gap}, where $\mat{T}^{\rm OT}$ is a solution of (\ref{eq:FormulationOptimalTransport}), i.e., $\mat{T}^{\rm OT}:=\defargmin \langle \mat{C}, \mat{T}\rangle$.
\end{definition}
The $\epsilon$-approximation is also defined. 
\begin{definition}[$\epsilon$-approximation w.r.t. evaluation function]
Consider an evaluation function $\phi: \mathbb{R}^{n \times n} \times \mathbb{R}^{n \times n} \to \mathbb{R}$ or $\phi: \mathbb{R}^{n \times n} \times \mathbb{R}^{n} \to \mathbb{R}$. For any $\epsilon > 0$, {the matrix $\mat{P}$} is called $\epsilon$-approximation matrix with respect to the function if $\phi({\mat{P},Q}) \leq  \epsilon$, where {$Q$} represents a matrix or vector. In addition, $\epsilon$ is called the approximation constant.
{When $\phi (\mat{T},\hat{\mat{T}}) = f(\mat{T}) - f(\hat{\mat{T}})$, we designate it as the $\epsilon$-approximation with respect to the functional value {gap}, where $\hat{\mat{T}}$ is a matrix of (\ref{eq:KLSemiRelaxed}), i.e., $\hat{\mat{T}}:=\defargmin f(\mat{T})$. In case of $\phi (\mat{T},\vec{a}) = \|\mat{T}\vec{1}_n - \vec{a}\|$ where $\vec{a}$ is an input vector, $\phi$ is called the $\epsilon$-approximation of the marginal constraint {gap}. Similarly, $\phi (\mat{Y},\mat{T}^{\rm OT}) = \langle \mat{C} , \mat{Y} \rangle - \langle \mat{C} , \mat{T}^{\rm OT} \rangle$ is {called} the $\epsilon$-approximation of the OT distance gap. }
\end{definition}

\subsection{Convergence analysis of functional value {gap} based on new proof strategy }
\label{Sec:FuncAnalysis}

We begin to give the convergence to $\epsilon$-approximation in terms of the functional value {gap}.
\begin{Thm}[convergence to $\epsilon$-approximation {w.r.t.} functional value {gap}]
\label{thm:AlgorithmTimeComplexity}
Letting $c_1$ and $c_2$ respectively represent {$(\frac{2n(\tau+\eta)R}{\tau} + 1) \beta$} and $2\beta \log n$, then one can consider the case in which $k$ is even after the odd update. If $\|\log \left ( \frac{{\bf T}^{(k)}}{{\bf T}^{*}}\right) \|_\infty \leq \epsilon^{\prime}$, $\mat{T}^{(k)}$ generated by the SR--Sinkhorn algorithm satisfies
\begin{equation*}
	f(\mat{T}^{(k)})-f(\hat{\mat{T}}) \leq (\beta \|\mat{C}\|_1 + \tau c_1)\epsilon^{\prime}+ \eta c_2.
\end{equation*}
Furthermore, defining $\epsilon^{\prime} = \frac{\epsilon_f}{2(\beta\|{\bf {C}}\|_1 +\tau c_1)}$ and $\eta = \frac{\epsilon_f}{2c_2}$ for an approximation constant $\epsilon_f$, $f(\mat{T}^{(k)})-f(\hat{\mat{T}}) \leq \epsilon_f$ holds.
\end{Thm}

This theorem engenders the following corollary about the total complexity.
\begin{Cor}[stopping iteration bound and total complexity]
\label{cor:ComplexityAndOrder}
Letting $c_1$ and $c_2$ respectively denote {$(\frac{2n(\tau+\eta)R}{\tau} + 1) \beta$}  and $2\beta \log n$, then {\bf Theorem \ref{thm:AlgorithmTimeComplexity}} holds. The stopping iteration bound is given as
\begin{equation*}
	 k \geq 2  \Bigl(1+\frac{2c_2\tau}{\epsilon_f}\Bigr) \Bigl(\log 16\tau R  + \log c_2(\beta\|\mat{C}\|_1+ \tau c_1)  + 2\log \frac{1}{\epsilon_f} \Bigr) + 3.
\end{equation*}
Furthermore, assuming $R = \mathcal{O}(\frac{1}{\eta}\|\mat{C}\|_\infty)$, the total complexity of the SR--Sinkhorn algorithm  is 
\begin{equation*}
	\mathcal{O}\left ( \frac{\tau n^2}{\epsilon_f} \log n  \left (  \log n + \log \tau + \log (\log n) + \log \|\mat{C}\|_\infty + {\log (n^2  + \frac{n \tau \log n}{\epsilon_f})}  + \log \frac{1}{\epsilon_f} \right ) \right).
\end{equation*}
\end{Cor}

For the proof of {\bf Theorem \ref{thm:AlgorithmTimeComplexity}}, we first give three necessary lemmas without the proofs. The proofs of these lemmas are presented in the supplementary material. 
\begin{Lem}
\label{lem:NewStoppingCreterion}
When the iteration $k$ is even, the $k$-th matrix $\mat{T}^{(k)}$ satisfies the following inequality.
\begin{equation*}
\left|\!\left|\log \left ( \frac{\mat{T}^{(k)}}{\mat{T}^{*}}\right) \right|\!\right|_\infty \leq \frac{4\tau}{\eta}R \left (\frac{\tau}{\tau + \eta}\right )^{\frac{{k-1}}{2}-1},
\end{equation*}
where $R = \max \lbrace \| \log (\vec{a} )\|\infty , \| \log(\vec{b})\|\infty \rbrace + \max  \left\{ \log(n),\frac{1}{\eta}\| \mat{C}\|_\infty - \log(n) \right\}.$
\end{Lem}
\begin{Lem}
\label{lem:LogarithmInequality}
For {$0 < y < x < b$},  the logarithm function satisfies the following .
\begin{equation*}
	\log x - \log y\geq \frac{1}{b}{(x - y)}.
\end{equation*} 
\end{Lem}
\begin{Lem}{}
\label{lem:KeyInequality}
The iteration $k$ is even after the odd update. The optimal solution $\mat{T}^*$ and the $k$-th iterate $\mat{T}^{(k)}$ generated by the SR--Sinkhorn algorithm satisfy the following inequality.
\begin{equation*}
	\|\log \mat{T}^{(k)} - \log \mat{T}^*\|_\infty \geq \frac{1}{\beta} \|\mat{T}^{(k)} - \mat{T}^* \|_\infty.
\end{equation*}
\end{Lem}
It should be emphasized that {\bf Lemma \ref{lem:LogarithmInequality}} derives the important lemma {\bf Lemma \ref{lem:KeyInequality}}. In addition, these lemmas play crucially important roles in the proof of {\bf Theorems \ref{thm:MarginalConstraintError}} and {\bf  \ref{thm:approximation_originalOT}}. 
We now give the proof of {\bf Theorem \ref{thm:AlgorithmTimeComplexity}}.
\begin{proof}
We assume that the number iteration $k$ is even. Then, we have the followings.
\begin{align*}
f(\mat{T}^{(k)})-f(\hat{\mat{T}})&= g(\mat{T}^{(k)})-g(\hat{\mat{T}})+\eta(\mathrm{H}(\mat{T}^{(k)})-\mathrm{H}(\hat{\mat{T}}))\\
&= g(\mat{T}^{(k)})-g(\hat{\mat{T}})+g(\mat{T}^*)-g(\mat{T}^*)+\eta(\mathrm{H}(\mat{T}^{(k)})-\mathrm{H}(\hat{\mat{T}}))\\
&\leq g(\mat{T}^{(k)})-g(\mat{T}^*)+\eta(\mathrm{H}(\mat{T}^{(k)})-\mathrm{H}(\hat{\mat{T}}))\\
&\leq \langle  \mat{C},\mat{T}^{(k)} - \mat{T}^* \rangle + \tau (\mathrm{KL}(\mat{T}^{(k)}\vec{1}_n,\vec{a}) - \mathrm{KL}(\mat{T}^*\vec{1}_n,\vec{a}))+\eta(\mathrm{H}(\mat{T}^{*})-\mathrm{H}(\hat{\mat{T}})),
\end{align*}
where the first inequality uses $ g(\mat{T}^*)\leq g(\hat{\mat{T}})$ because $\mat{T}^*$ is the optimal solution of $g(\mat{T})$. The term $\langle \mat{C}, \mat{T}^{(k)} - \mat{T}^* \rangle$ is bounded by the Holder's inequality as
\begin{eqnarray*}
	\langle  \mat{C} , \mat{T}^{(k)} - \mat{T}^* \rangle \leq  \|\mat{C}\|_1\|\mat{T}^{(k)} - \mat{T}^* \|_\infty \leq \beta \|\mat{C}\|_1\| \log\mat{T}^{(k)} - \log\mat{T}^*\|_\infty,
\end{eqnarray*}
where the second inequality uses {\bf Lemma \ref{lem:KeyInequality}}. Here, we consider the stopping criterion in {\bf Lemma \ref{lem:NewStoppingCreterion}}. Consequently, the term is bounded by
\begin{equation}
\label{eq:bound_1}
\langle \mat{C} ,\mat{T}^{(k)} - \mat{T}^*\rangle \leq \beta \|\mat{C}\|_1\epsilon^{\prime}.
\end{equation}
From {the full proof} in the supplementary material, we can bound the KL and entropy terms as $\epsilon^{\prime}c_1$ and $c_2$ respectively, where {$c_1 = (\frac{2n(\tau+\eta)R}{\tau} + 1) \beta$}  and $c_2 = 2\beta \log n$. Finally, putting all of them together yields
\begin{eqnarray*}
	f(\mat{T}^{(k)})-f(\hat{\mat{T}}) &\leq& \beta \|\mat{C}\|_1\epsilon^{\prime} + \tau c_1 \epsilon^{\prime}+ \eta c_2.
\end{eqnarray*}
Setting $\eta = \frac{\epsilon_f}{2c_2}, \epsilon^{\prime} \!=\! \frac{\epsilon_f}{2(\beta\|{\bf {C}}\|_1+\tau c_1)}$, $f(\mat{T}^{(k)})-\!f(\hat{\mat{T}}) $ is bounded by the approximation constant $\epsilon_f$. This completes the proof.
\end{proof}

\begin{Remark}
The obtained results resemble those of \cite{NeurIPS2021robust}. However, this is obtained by constructing a new proof strategy. Our proof particularly addresses the {\it upper bound of transport matrix} instead of the properties of the functional values as adopted in \cite{NeurIPS2021robust}. More specifically, the UOT--Sinkhorn and RS--Sinkhorn algorithms respectively address the property of the functional values as Lemma 4 of \cite{UOTSinkhorn2020} and the equality (29) of the supplementary \cite{NeurIPS2021robust}. Therefore, they address neither the inequalities bounding the differences $\mat{T}^{(k)} \!-\! \mat{T}^*$, nor the bound {of} $\langle \mat{C},\mat{T}^{(k)} \!-\! \mat{T}^* \rangle$. However, deriving and using {\bf Lemma \ref{lem:LogarithmInequality}}, we do not only evaluate the distance between $\mat{T}^{(k)}$ and $\mat{T}^*$ directly, but also guarantee the marginal constraint {gap} as well as {the $\epsilon$-approximation of {the OT distance gap}}.  
\end{Remark}

\subsection{Convergence analysis of {the} marginal constraint {gap}}
\label{Sec:MarginalAnalysis}

We first provide the convergence rates in the marginal constraint {gap}s of not only the non-relaxed constraint of the vector $\vec{b}$ but also the relaxed constraint of $\vec{a}$.

\begin{Thm}[convergence rates of marginal constraint gap]
\label{thm:MarginalConstraintError}
One can consider the case for even $k$ after the odd update. Then, the marginal constraint {gap} of the vector $\vec{a}$ satisfies
\begin{eqnarray}
	\label{eq:marginal_bound_a}
	\|\vec{a}^{(k)} - \vec{a}\|_\infty \leq \gamma \left ( \frac{4\tau}{\eta}R\left (\frac{\tau}{\tau+\eta}\right )^{ \frac{{k-1}}{2}  -1}  + \frac{\|\vec{u}^*\|_\infty}{\tau}\right),
\end{eqnarray}
where $\gamma = \max \lbrace \alpha,\beta \rbrace$ and $\vec{u}^*$ is the optimal solution of the dual form of (\ref{eq:EntropySemiRelaxed}). Furthermore, assuming that $k$ is odd after the even update, the logarithm marginal gap of $\vec{b}$ is bounded by
\begin{equation}
\label{eq:marginal_bound_b}
	\| \log(\vec{b}^{(k)}) - \log(\vec{b}) \|_\infty \leq \frac{4\tau}{\eta}R\left (\frac{\tau}{\tau+\eta}\right )^{{\frac{k-1}{2}} -1} .
\end{equation}
\end{Thm}
\begin{proof}
We can provide a relevant proof sketch. Because $\vec{a}^{(k)}_i \leq \beta$ and $\vec{a}_i \leq \alpha$ hold, we use {\bf Lemma \ref{lem:LogarithmInequality}} and transform $\| \vec{a}^{(k)} - \vec{a} \|_\infty$ into $\gamma \|\log \vec{a}^{(k)} - \log \vec{a}\|_\infty$. Then, applying the inequality about geometric convergence, we obtain the inequality (\ref{eq:marginal_bound_a}). Similarly, we derive the inequality (\ref{eq:marginal_bound_b}).
\end{proof}

\begin{Remark}
From (\ref{eq:marginal_bound_a}), we understand how the marginal constraint {gap} converges to the second term of the right-side of (\ref{eq:marginal_bound_a}). If {\bf Lemma \ref{lem:NewStoppingCreterion}} holds, we obtain the marginal constraint {gap} that can be kept within the approximation constant $\epsilon$ towards the second term. It is noteworthy that, because of the inconsistency between $\alpha$ and $\beta$ in the SROT problem, the second term of (\ref{eq:marginal_bound_a}) does not go to zero.
\end{Remark}

Fortunately, additionally assuming that two distributions $\vec{a}$ and $\vec{b}$ are the probability simplex $\Delta^n$, we are able to bound the second term of the inequality (\ref{eq:marginal_bound_a}) by given parameters. We {then} finally obtain the convergence rate to the $\epsilon$-approximation in terms of the marginal constraint {gap}. This modification has no effect on the algorithm. We provide this theorem below.

\begin{Thm}[convergence to $\epsilon$-approximation of marginal constraint {gap}]
\label{thm:AlgorithmTimeMarginalComplexity}
Assume $\vec{a},\vec{b} \in \Delta^n$. Consider the case in which $k$ is even after the odd update.
Then, the marginal constraint {gap} of the vector $\vec{a}$ can be newly bounded as
\begin{equation}
	\label{eq:new_marginal_bound_a}
	\|\vec{a}^{(k)} - \vec{a}\|_\infty \leq \frac{4\tau}{\eta}R\left (\frac{\tau}{\tau + \eta}\right )^{ \frac{{k-1}}{2}  -1}  + {\frac{U}{\tau+\eta}},
\end{equation}
where $\vec{a}_{\rm max}$ and $\vec{a}_{\rm min}$ denote the maximum and minimum element of $\vec{a}$, respectively. $U$ represents $U =  \|\mat{C}\|_\infty+\eta \log \left( \vec{a}_{\rm max}/\vec{a}_{\rm min}\right)$.
Also if $\|\log \left ( \frac{{\bf T}^{(k)}}{{\bf T}^{*}}\right) \|_\infty \leq \epsilon^{\prime}$, the bound inequality of the marginal constraint gap is given as
\begin{equation}
\label{eq:new_marginal_bound_b}
	\|\vec{a}^{(k)} - \vec{a}\|_\infty \leq \epsilon^\prime + {\frac{U}{\tau+\eta}}.
\end{equation}
{Furthermore, for $\epsilon_c \leq 2 \log \left ( {\vec{a}_{\rm max}}/{\vec{a}_{\rm min}} \right)$, taking $\tau = ({2\|\mat{C}\|_\infty})/{\epsilon_c} + \eta(\frac{2}{\epsilon_c}\log \left ( {\vec{a}_{\rm max}}/{\vec{a}_{\rm min}} \right) - 1), \forall {\eta >0}$, $\epsilon^\prime = \frac{\epsilon_c}{2}$, $\|\vec{a}^{(k)} - \vec{a}^* \|_\infty \leq \epsilon_c$ holds. Otherwise, for $\epsilon_c \geq 2 \log \left ( {\vec{a}_{\rm max}}/{\vec{a}_{\rm min}} \right)$, taking $\tau = (2\|\mat{C}\|_\infty)/\epsilon_c + \eta(\frac{2}{\epsilon_c}\log \left ( {\vec{a}_{\rm max}}/{\vec{a}_{\rm min}} \right) - 1), \forall \eta, 0 \leq \eta \leq (2\|\mat{C}\|_\infty) / (\epsilon_c(1 - \frac{2}{\epsilon_c}\log \left ( {\vec{a}_{\rm max}}/{\vec{a}_{\rm min}} \right) ))$, and $\epsilon^\prime = \frac{2}{\epsilon_c}$, then $\|\vec{a}^{(k)} - \vec{a}^* \|_\infty \leq \epsilon_c$ holds.}
\end{Thm}

\begin{Remark}
The upper bound of the second term is, in general, greater than or equal $1$ because $\|\vec{u}\|_\infty$ is bounded as $2(\tau+\eta)R$ in {\bf Lemma \ref{lem:SupNormOptimalInqquality}} in the supplementary material. Therefore, we cannot bound the relaxed marginal constraint {gap} for an arbitrary constant. However, when $\vec{a},\vec{b} \in \Delta^n$, we can diminish the value of the second term using the parameters $\tau$ and $\eta$. From this fact, given a marginal approximation constant $\epsilon$, we can calculate the necessary number of iterations such that the marginal constraint {gap} falls below $\epsilon$. 
\end{Remark}

\subsection{Convergence analysis of {the} OT distance gap}
\label{Sec:OTDistanceAnalysis}

{This subsection provides the $\epsilon$-approximation of the OT distance gap. On the condition that the SROT problem has the probability simplex constraints of $\vec{a}$ and $\vec{b}$, we further consider to project the final $\mat{T}^{(k)}$ generated by the SR--Sinkhorn algorithm onto $\mathcal{U}(\vec{a}, \vec{b})$ in (\ref{eq:TransportPolytope}) using the projection operator \cite[Algorithm 2]{altschuler2017nearlinear} to measure the gap. Under this setting, we derive the following result.}

\begin{Thm}[convergence to $\epsilon$-approximation of OT distance gap]
\label{thm:approximation_originalOT}
Letting $\mat{T}^{\rm OT}$, $c_3$ and $U$ be the optimal solution of {the standard OT problem (\ref{eq:FormulationOptimalTransport})}, $2\log n  + 1 - \defmax \lbrace \mathrm{H}(\vec{a}), \mathrm{H}(\vec{b}) \rbrace)$ and {$ \|\mat{C}\|_\infty+\eta \log \left( {\vec{a}_{\rm max}}/{\vec{a}_{\rm min}}\right)$, respectively}. Assume $\vec{a},\vec{b} \in \Delta^{n}$. Also, one considers the case in which $k$ is even after the odd update. Let $\mat{T}^{(k)}$ and $\mat{Y}$ be the matrix generated by the SR--Sinkhorn algorithm and {its projected matrix by \cite[Algorithm 2]{altschuler2017nearlinear}, respectively}. Then, the OT distance gap is provided as
\begin{equation}
\label{eq:New_Distance_Bound}
	\langle \mat{C},\mat{Y} \rangle - \langle \mat{C},\mat{T}^{\mathrm{OT}} \rangle \leq (2n\|\mat{C}\|_{\infty}+\|\mat{C}\|_1)\frac{4\tau}{\eta}R\left (\frac{\tau}{\tau + \eta}\right )^{ \frac{{k-1}}{2}  -1} +\eta c_3 + \frac{2n\|\mat{C}\|_{\infty}}{\tau}U.
\end{equation}
In addition, if $\|\log \left ( \frac{{\bf T}^{(k)}}{{\bf T}^{*}} \right) \|_\infty \leq \epsilon^{\prime}$, {$\mat{Y}$} satisfies
\begin{equation}
\label{eq:approximationOTdistance}
	\langle \mat{C},{\mat{Y}} \rangle - \langle \mat{C}, \mat{T}^{\rm OT} \rangle \leq (2n\|\mat{C}\|_{\infty}+\|\mat{C}\|_1)  \epsilon^\prime  + \eta c_3 + \frac{2n\|\mat{C}\|_\infty}{\tau}U.
\end{equation}
Defining $\epsilon^\prime = \frac{\epsilon_d}{{3(2n\|{\bf C}\|_\infty + \|{\bf C}\|_1)}}$, $\eta = \frac{\epsilon_d}{3c_3}$ and $\tau = \frac{6n\|{\bf C}\|_\infty}{{\epsilon_d}}U$, $\langle \mat{C},\mat{Y} \rangle - \langle \mat{C}, \mat{T}^{\rm OT} \rangle \leq \epsilon_d$ holds.
\end{Thm}
\begin{proof}
The proof sketch can be provided. For $\langle \mat{C}, \mat{Y} - \mat{T}^{\rm OT} \rangle$, we have
\begin{eqnarray*}
	\langle \mat{C},{\mat{Y}} - \mat{T}^{\rm OT} \rangle  &=& {\langle \mat{C}, \mat{Y} - \mat{T}^{(k)} \rangle} + \langle \mat{C},\mat{T}^{(k)} - \mat{T}^{*} \rangle  + \langle \mat{C}, \mat{T}^{*} -  \mat{T}^{\rm OT} \rangle \\
	&\leq&  \|\mat{C}\|_\infty \|\mat{Y} - \mat{T}^{(k)}\|_1+ \|\mat{C}\|_1\|\mat{T}^{(k)} - \mat{T}^{*}\|_\infty  + \langle \mat{C}, \mat{T}^{*} -  \mat{T}^{\rm OT} \rangle.
\end{eqnarray*}
{From \cite[Lemma 7]{altschuler2017nearlinear} and the upper bound of {\bf Theorem \ref{thm:AlgorithmTimeMarginalComplexity}}, we bound $\|\mat{Y} - \mat{T}^{(k)}\|_1$ as $\frac{2n}{\tau}U$}. Additionally, {\bf Lemma \ref{lem:KeyInequality}} bounds $\|\mat{T}^{(k)} - \mat{T}^{*}\|_\infty$ as $\|\mat{C}\|_1\epsilon^\prime$. Also, we bound $\langle \mat{C}, \mat{T}^{*} -  \mat{T}^{\rm OT} \rangle$ as $\eta c_3$. Then, taking $\epsilon^\prime = \frac{\epsilon_d}{{3(2n\|{\bf C}\|_\infty + \|{\bf C}\|_1)}}$, $\eta = \frac{\epsilon_d}{2c_3}$ and $\tau = \frac{6n\|{\bf C}\|_\infty}{\epsilon}U$, {\bf Theorem \ref{thm:approximation_originalOT}} holds.
\end{proof}

\begin{Remark}
As this proof sketch reveals, {\bf Lemma \ref{lem:LogarithmInequality}} is necessary to derive the $\epsilon$-approximation of {the OT distance gap. It should be also noted that {\bf Theorem \ref{thm:AlgorithmTimeMarginalComplexity}} is a must in deriving this result.}
\end{Remark}

\section{Numerical analysis}
\label{sec:NumericalAnalysis}

This section evaluates the theoretical marginal constraint gap and the OT distance gap in comparison with the empirical results. {Additional experiments are in the supplementary material.}

\subsection{Evaluation of the marginal constraint gap.}
We evaluate {\bf Theorem \ref{thm:AlgorithmTimeMarginalComplexity}} {using} a synthetic dataset. We uniformly generate the {ground} cost matrix $\mat{C}$ from the closed interval $[1,10]$ with $n=50$. The weight vectors $\vec{a}$ and $\vec{b}$ are configured uniformly from the closed interval $[1,5]$, and are normalized to $1$. We select $(\tau,\eta) = (10^6,10^{-2})$. The maximum iteration number is $2000$. We used Python Optimal Transport library\footnote{\url{https://pythonot.github.io/}.} \cite{flamary2021pot}  and the UOT code\footnote{\url{https://github.com/lntk/uot}.}.
We compare the empirical result of $\|\mat{T}^{(k)}\vec{1}_n - \vec{a}\|_\infty$ with the theoretical upper bound at optimum using the second term in (\ref{eq:new_marginal_bound_a}) because the first term disappears at that point. Figure \ref{fig:TheoreticalGapComparison}(a) portrays these two values at the iterations after the odd update because {\bf Theorem \ref{thm:AlgorithmTimeMarginalComplexity}} holds when iterations are even. {Thereby,} the $x$-axis of the figure ranges from $0$ to $1000$.
{From this figure, we see that theoretical upper bound gets close to the empirical result, where both of them are sufficiently close to zero. Consequently, we can understand that the obtained bound is tight,} and this result supports {\bf Theorem \ref{thm:AlgorithmTimeMarginalComplexity}}.

\subsection{Evaluations of the OT distance gap.}
\begin{figure}[t]
\begin{center}
	\begin{minipage}[t]{0.49\textwidth}
	\begin{center}
	\includegraphics[width=0.99\textwidth]{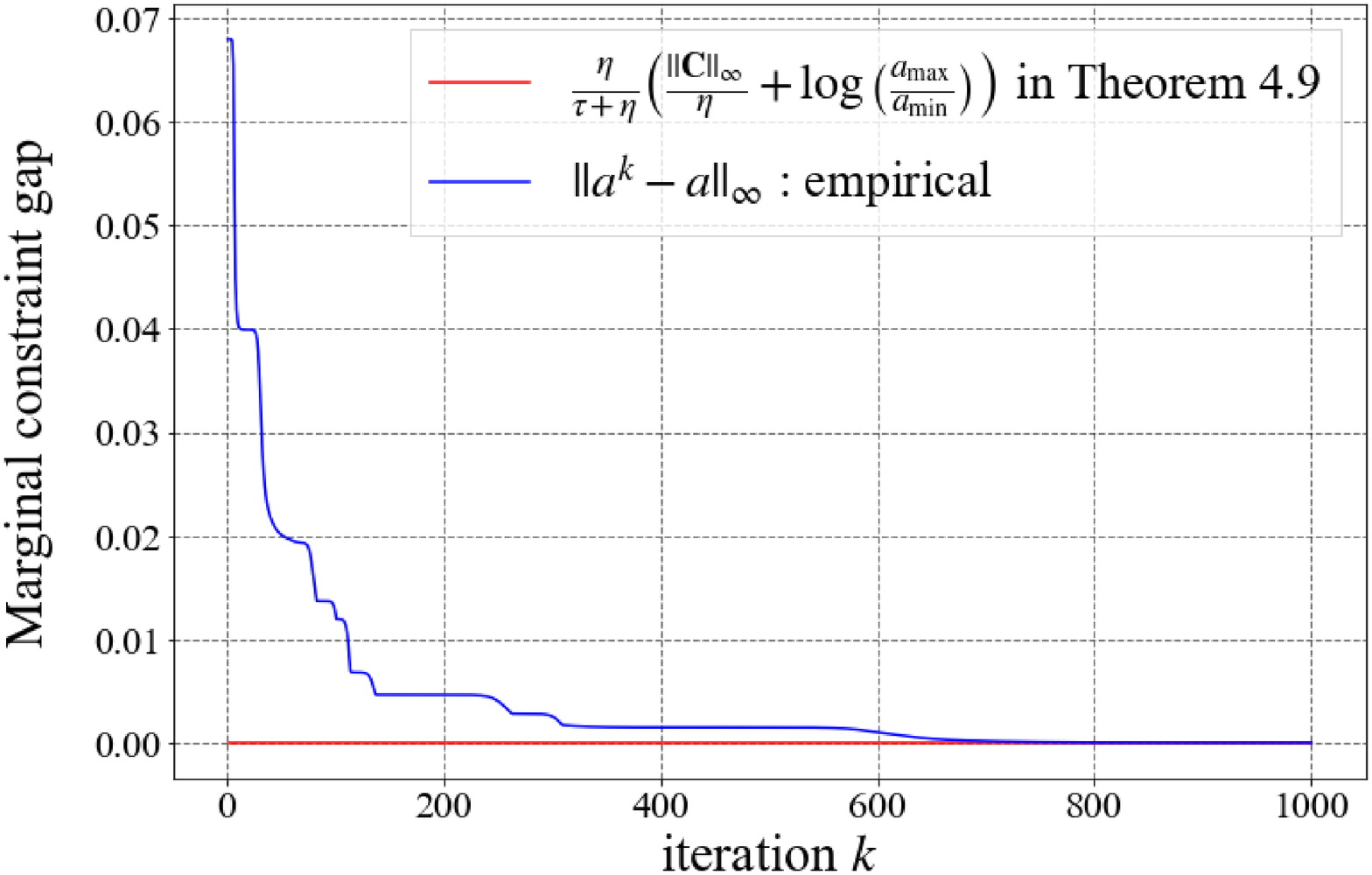}	
	
	{\small (a) $\|\mat{T}^{(k)}\vec{1}_n - \vec{a}\|_{\infty}$.}
	\end{center}
	\end{minipage}
	\begin{minipage}[t]{0.49\textwidth}
	\begin{center}	
	\includegraphics[width=0.99\textwidth]{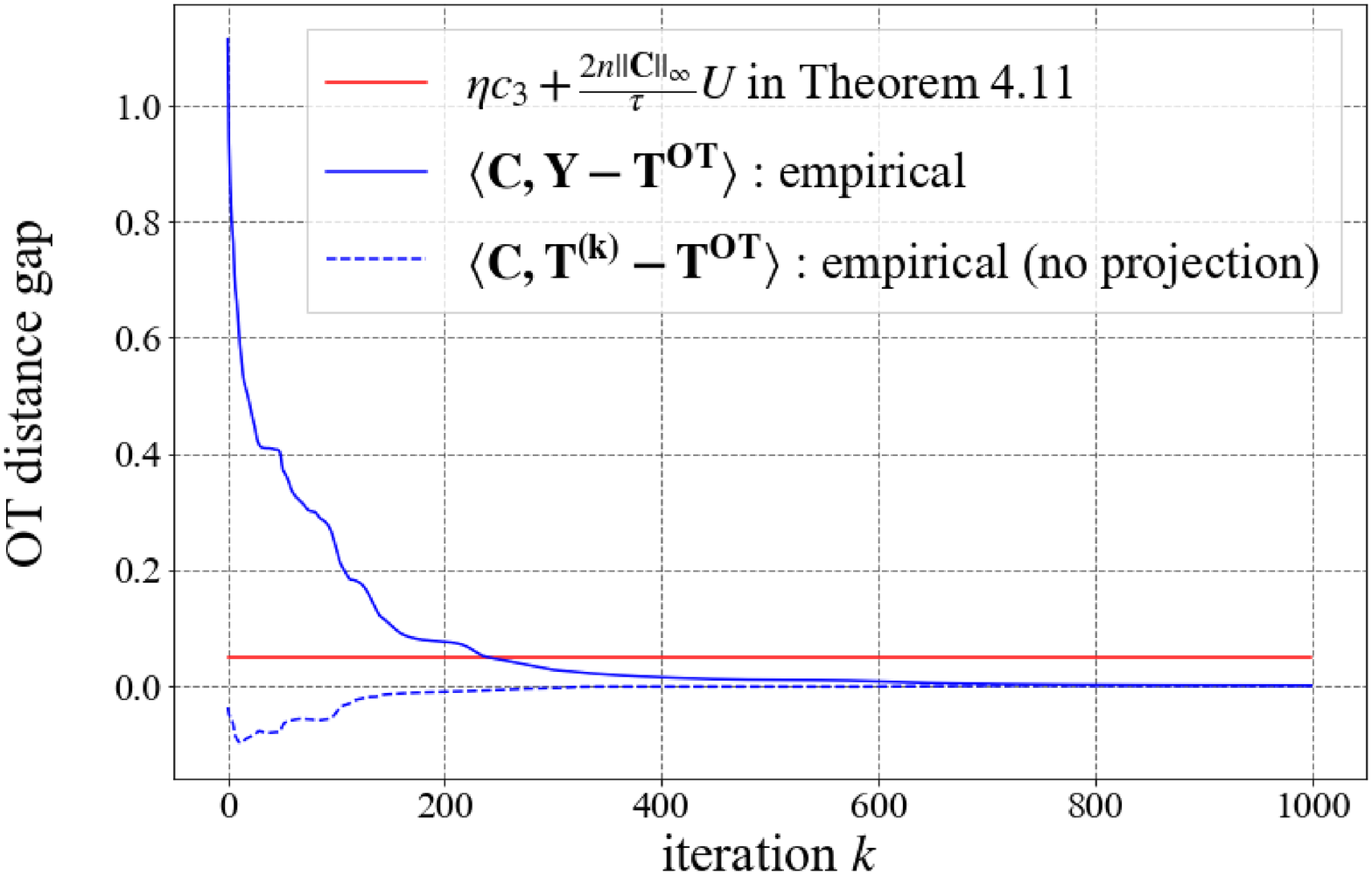}	
	
	{\small (b) $\langle \mat{C},\mat{Y} - \mat{T}^{\mathrm{OT}} \rangle$.}
	\end{center}	
	\end{minipage}	
	\caption{Left: Marginal constraint gap. Blue line represents empirical results, and red line is {the} theoretical upper bound in (\ref{eq:new_marginal_bound_a}). Right: OT distance gap. Blue solid and dashed lines represent the empirical OT distance gap and the non-projected one, respectively. Red line is {the} theoretical upper bound in (\ref{eq:New_Distance_Bound}).}	
	\label{fig:TheoreticalGapComparison}	
\end{center}		
\end{figure}

We next evaluate {\bf Theorem \ref{thm:approximation_originalOT}}. The synthetic dataset is the same as that of the previous experiment. 
Similarly to the previous experiment, we compare the empirical result of $\langle \mat{C},\mat{Y} \rangle - \langle \mat{C},\mat{T}^{\mathrm{OT}}\rangle$ with the theoretical upper bound at optimum based on the second and third terms in (\ref{eq:New_Distance_Bound}). The OT distance gap without the projection is also measured as reference. From Figure \ref{fig:TheoreticalGapComparison}(b), we see that the two empirical results go to zero as the iteration increases.  Interestingly, even the non-projected OT distance gap approaches zero. Although the theoretical upper bound does not approach zero under this parameter setting, it is still close to zero. We expect that the larger $\tau$ with smaller $\eta$ will get closer to zero. Therefore, this result verifies {\bf Theorem \ref{thm:approximation_originalOT}}.

\section{Conclusion}
This paper has presented a comprehensive convergence analysis of the SR--Sinkhorn algorithm for the semi-relaxed optimal transport (SROT) problem. It is noteworthy that we have newly provided the upper bound of the marginal constraint gap exploiting our new proof strategy. We also provided its $\epsilon$-approximation when two distributions are in the probability simplex. Moreover, the convergence analysis of the OT distance gap to the $\epsilon$-approximation is given with the help of the obtained marginal constraint gap. Our future work is to provide $\epsilon$-approximation of the Gromov Wasserstein distance through its relaxed variants.

\clearpage

\bibliography{reference,kasailab}
\bibliographystyle{IEEEtranSN}

\clearpage

\appendix

\begin{center}
{\bf\Large Supplementary}
\end{center}

\renewcommand\thefigure{A.\arabic{figure}}  
\setcounter{figure}{0} 

\renewcommand\thetable{A.\arabic{table}}  
\setcounter{table}{0} 

\renewcommand{\theequation}{A.\arabic{equation}}
\setcounter{equation}{0}

\renewcommand\thealgorithm{A.\arabic{algorithm}}  
\setcounter{algorithm}{0} 

\renewcommand\thefigure{A.\arabic{figure}}  
\setcounter{figure}{0} 

\renewcommand\thetable{A.\arabic{table}}  
\setcounter{table}{0} 

\renewcommand{\theequation}{A.\arabic{equation}}
\setcounter{equation}{0}

\renewcommand\thealgorithm{A.\arabic{algorithm}}  
\setcounter{algorithm}{0}

This supplementary material presents the deviation of the dual form of the entropy regularized SROT, the description of SR-Sinkhorn, the complete proof of theoretical result that are provided in the main material, and additional experiments. The structure is as follows:
\begin{itemize}
\item {\bf Section A:}
\newline
The deviation of the dual form for the entropy regularized SROT and the algorithmic description and pseudocode of the SR-Sinkhorn algorithm.
\item {\bf Section B:}
 \newline
The complete proof of {\bf Theorem \ref{thm:AlgorithmTimeComplexity}}.
\item {\bf Section C:}
 \newline
The complete proofs of {\bf Theorems \ref{thm:MarginalConstraintError}} and {\bf \ref{thm:AlgorithmTimeMarginalComplexity}}.
\item {\bf Section D:}
 \newline
The complete proof of {\bf Theorem \ref{thm:approximation_originalOT}}.
\item {\bf Section E:}
\newline
We newly add the theoretical result of the marginal constraint gap bound of the SROT problem {without the entropy regularization, which is defined in  (\ref{eq:KLSemiRelaxed})}. This result is not included in the main material.
\item {\bf Section F:}
\newline
Additional experimental results. 
\end{itemize}

\section{Entropy regularized semi-relaxed Sinkhorn algorithm: {\scshape SR--Sinkhorn}}
This section presents the deviation of the {\it Fenchel dual} of the entropy regularized SROT and SR--Sinkhorn. Note that their similar formulation and algorithm are proposed in \cite{NeurIPS2021robust}.

\subsection{The dual form of entropy regularized SROT}
\label{sec:FormulationEntropySemiRelaxed}

{For completeness, this subsection explains the derivation} of a dual problem to construct our proposed algorithm. We consider the Fenchel dual of the primal problem in (\ref{eq:EntropySemiRelaxed}). We give the dual objective as
\begin{eqnarray*}
\min_{\substack{{\bf T}^T\bm{1}_n = \bm{b}\\ {\bm{y}} \in \mathbb{R}^n_+}} \langle\mat{C},\mat{T} \rangle + \tau\mathrm{KL}(\vec{y},\vec{a})-\eta \mathrm{H}(\mat{T})+\vec{u}^T(\vec{y}-\mat{T}\vec{1}_n),
\end{eqnarray*}
where $\bm{u} \in \mathbb{R}^n$. This problem is separable as the two following minimum problems.
\begin{eqnarray}
\label{eq:FirstObjective}
\min_{{\bf T} \geq \bm{0},{\bf T}^T\bm{1}_n = \bm{b}} &\!\!\!\!\!\!\!\!&
\langle\mat{C},\mat{T} \rangle-\vec{u}^T\mat{T}\vec{1}_n-\eta\mathrm{H}(\mat{T}),\\
\label{eq:SecondObjective}
\min_{\bm{y} \in \mathbb{R}^n_+} &\!\!\!\!\!\!\!\!&
\vec{u}^T\vec{y} +\tau\mathrm{KL}(\vec{y},\vec{a}).
\end{eqnarray}
The first problem (\ref{eq:FirstObjective}) is solvable using the KKT conditions. Therefore, introducing a dual variable $\vec{v} \in \mathbb{R}^n$, we derive the corresponding dual problem given as shown below.
\begin{equation*}
	{\max_{\bm{v} \in \mathbb{R}^n}}\  -\eta \sum_{i,j}\exp \left( \frac{\vec{u}_i+\vec{v}_j - \mat{C}_{i,j}}{\eta} \right) +\vec{v}^T\vec{b}.
\end{equation*}
The second problem (\ref{eq:SecondObjective}) is convex and the minimum can be found considering the point where the gradient is zero vector. This yields
\begin{equation*}
	 {\max_{\bm{u} \in \mathbb{R}^n}}\  -\tau \left( \vec{a}^T\exp \left( -\frac{\vec{u}}{\tau} \right) - \vec{a}^T\vec{1}_n\right).
\end{equation*}
Combining the two optimization problems yields the dual maximization problem for $(\vec{u},\vec{v})$. By flipping the sign of the problem above, one can formulate the final dual minimization problem as presented below.
\begin{equation}
\label{eq:FenchelDualSemiRelaxed}
\min_{\bm{u},\bm{v}}\ \eta\sum_{i,j}\exp \left(\!\frac{\vec{u}_i+\vec{v}_j - \mat{C}_{i,j}}{\eta} \!\right) - \vec{v}^T\vec{b} +\tau \vec{a}^T\!\exp \left(\! -\frac{\vec{u}}{\tau}\! \right).
\end{equation}
Now, we designate this internal terms as $h(\vec{u},\vec{v})$.

\subsection{Semi-relaxed Sinkhorn algorithm: {\scshape SR--Sinkhorn}}
\label{sec:ProposalSinkhornAlgorithm}
We present an alternative optimization algorithm to find the optimal solution of the dual problem (\ref{eq:FenchelDualSemiRelaxed}). Because the function $h(\vec{u},\vec{v})$ is biconvex for $(\vec{u},\vec{v})$, the alternative optimization algorithm can get the global optimal solution. We attempt to solve the solution at iteration $k+1$, using the $k$-th iteration solution $(\vec{u}^{(k)},\vec{v}^{(k)})$. 
When considering $\vec{u}^{(k+1)}_i$ under the fixed $\vec{v}^{(k)}$, calculating the gradient of $h$ for $(\vec{u}^{(k+1)}_i, \vec{v}^{(k)})$ satisfies
\begin{equation*}
	\vec{a}_i\exp\left(-\frac{\vec{u}^{(k+1)}_i}{\tau}\right)=\eta\sum_{j}\exp\left(\frac{\vec{u}^{(k+1)}_i+\vec{v}^{(k)}_j - \mat{C}_{ij}}{\eta}\right),
\end{equation*}
where the $k$-th transport matrix is defined as $\mat{T}^{(k)}_{i,j} = \exp(\frac{\bm{u}^{(k)}_i+\bm{v}^{(k)}_j - {\bf C}_{ij}}{\eta})$ and $\vec{a}^{(k)}$ is denoted as $\vec{a}^{(k)} = \mat{T}^{(k)}\vec{1}_n$. Multiplying $\exp (\frac{\bm{u}^{(k)}_i}{\eta})$ by both sides of the inequality above, the right-hand side is replaced with $\vec{a}^{(k)}_i$. Taking the logarithm of both sides yields the update of $\vec{u}^{(k+1)}_i$ as
\begin{equation}
\label{eq:SinkhornUpdateEven}
\vec{u}^{(k+1)}_i = \frac{\tau}{\tau+\eta}(\vec{u}^{(k)}_{{i}} + \eta(\log(\vec{a}_i) - \log(\vec{a}^{(k)}_i))).
\end{equation}
Similarly, we obtain the update of $\vec{v}^{(k+1)}_j$ as
\begin{equation}
\label{eq:SinkhornUpdateOdd}
	\vec{v}^{(k+1)}_j = \vec{v}^{(k)}_j + \eta (\log(\vec{b}_j) - \log(\vec{b}^{(k)}_j)),
\end{equation}
where $\vec{b}^{(k)} = (\mat{T}^{(k)})^T\vec{1}_n$. 

It is noteworthy that the update of (\ref{eq:SinkhornUpdateOdd}) for odd $k$ {\it implicitly} represents the projection onto the primal constraints $\mat{T}^T\vec{1}_n = \vec{b}$. In fact, we have
\begin{eqnarray*}
\vec{b}^{(k+1)}_j &=& \sum_i \exp\left(\frac{\vec{u}^{(k+1)}_i+\vec{v}^{(k+1)}_j - \mat{C}_{ij}}{\eta}\right) \\
&=& \frac{\vec{b}_j}{\vec{b}^{(k)}_j}\sum_i \exp\left(\frac{\vec{u}^{(k)}_i + \vec{v}^{(k)}_j - \mat{C}_{ij}}{\eta}\right) 
= \frac{\vec{b}_j}{\vec{b}^{(k)}_j}\cdot \vec{b}^{(k)}_j = \vec{b}_j.
\end{eqnarray*}

{The derived SR--Sinkhorn algorithm is summarized in {\bf Algorithm \ref{Appenalg:AlternativeSemiRelaxedSinkhorn}}. Note that the derived SR--Sinkhorn algorithm is exactly the same as the Robust Semi Sinkhorn algorithm proposed in \cite{NeurIPS2021robust}}.
\begin{algorithm}[htbp]
\caption{Entropy Regularized Semi-Relaxed Sinkhorn Algorithm ({\scshape SR--Sinkhorn}) }      
\label{Appenalg:AlternativeSemiRelaxedSinkhorn}    
\begin{algorithmic}[1]
\REQUIRE{$\vec{a},\vec{b},\eta,\tau,\vec{u}^{(0)}=\vec{0},\vec{v}^{(0)} = \vec{0}$}
\ENSURE{$\vec{u}^{(k)},\vec{v}^{(k)}$}

\FOR {$k=0 \dots K$}
	\STATE {$\mat{T}^{(k)} = \mathrm{diag} (  \exp(\frac{\vec{u}^{(k)}}{\eta}) ) \exp(-\frac{\mat{C}}{\eta})\mathrm{diag} ( \exp(\frac{\vec{v}^{(k)}}{\eta}))$}
	\STATE {$\vec{a}^{(k)} = \mat{T}^{(k)}\vec{1}_n$ }
	\IF{$k$ is even}
	\STATE {$\vec{u}^{(k+1)} = \frac{\tau\eta}{\tau+\eta}(\frac{\vec{u}^{(k)}}{\eta} + \log(\vec{a}) - \log(\vec{a}^{(k)}))$}
		\STATE {$\vec{v}^{(k+1)} = \vec{v}^{(k)}$}
	\ELSE
		\STATE {$\vec{u}^{(k+1)} = \vec{u}^{(k)}$}
		\STATE {$\vec{v}^{(k+1)} = \vec{v}^{(k)} + \eta (\log(\vec{b}) - \log(\vec{b}^{(k)}))$}
	\ENDIF
\ENDFOR
\end{algorithmic}
\end{algorithm}

\section{Theoretical results about functional value gap}
This section presents the full proof of {\bf Theorem \ref{thm:AlgorithmTimeComplexity}}. We first provide the geometric convergence of the SR-Sinkhorn. Note that this geometric convergence has been already provided in \cite{NeurIPS2021robust}.
After that, we give and prove {three necessary lemmas, which are in the main material, to give} the proof of {\bf Theorem \ref{thm:AlgorithmTimeComplexity}}. 
Finally, we describe the full proof of {\bf Theorem \ref{thm:AlgorithmTimeComplexity}}. 

\subsection{Geometric Convegence on SR-Sinkhorn}
\begin{Thm}[convergence rate of dual solution gap ({\cite[{\bf Lemma 6}]{NeurIPS2021robust}})]
\label{thm:AlgorithmCnvergenceEnsure}
Let $(\vec{u}^*, \vec{v}^*)$ be the optimal solution of the dual problem in (\ref{eq:FenchelDualSemiRelaxed}). Then the solution $(\vec{u}^{(k+1)},\vec{v}^{(k+1)})$ generated by {\bf Algorithm \ref{Appenalg:AlternativeSemiRelaxedSinkhorn}} satisfies the following inequality
\begin{equation}
\label{eq:ConvergenceInequality}
\max \left \lbrace \| \vec{u}^{(k+1)}-\vec{u}^{*}\|_\infty,\|\vec{v}^{(k+1)}-\vec{v}^{*}\|_\infty \right \rbrace \notag  \leq 2\tau R\left (\frac{\tau}{\tau + \eta}\right )^{\frac{k}{2} -1},
\end{equation}
where $R$ is defined as 
\begin{equation*}
	R = \max \lbrace \| \log (\vec{a} )\|_{\infty} , \| \log(\vec{b})\|_{\infty} \rbrace + \max  \left\{ \log(n),\frac{1}{\eta}\| \mat{C}\|_\infty - \log(n) \right\}.
\end{equation*}
\end{Thm}
{To prove {\bf Theorem \ref{thm:AlgorithmCnvergenceEnsure}}, we first introduce two lemmas.} 
\begin{Lem}{}
\label{lem:OptimalEqualityDual}
	The optimal solution $\vec{u}^*$ of dual semi-relaxed problem, $\vec{a}^*$ and $\vec{b}^*$ satisfy the followings:	
	\begin{eqnarray*}
	\label{eq:OptimalEqualityRelaxed} 
	\frac{\vec{u}^*}{\tau} &=& \log \vec{a} - \log \vec{a}^*\\
	\label{eq:OptimalEqualityStrict} 
	\vec{b}^* & = &\vec{b}. 
	\end{eqnarray*}
\end{Lem}
\begin{proof}
The two equalities in this lemma are proved by considering the respective fixed points of the updates (\ref{eq:SinkhornUpdateEven}) and (\ref{eq:SinkhornUpdateOdd}). This completes the proof. 
\end{proof}

We next introduce {\bf Lemma \ref{lem:UnbalancedLemma2}}, which holds not only in case of unbalanced optimal transport (UOT) problem but also in case of a semi-relaxed OT (SROT) problem.
\begin{Lem}{}({\cite[{\bf Lemma 2}]{UOTSinkhorn2020}}) In {\bf Algorithm \ref{Appenalg:AlternativeSemiRelaxedSinkhorn}}, the two following inequalities are true.
\label{lem:UnbalancedLemma2}
\begin{eqnarray}
	\label{eq:UnbalancedLemmaA}
	\left|\log \frac{\vec{a}^*_i}{\vec{a}^{(k)}_i} - \frac{\vec{u}^*_i - \vec{u}^{(k)}_i}{\eta} \right| 
	&\leq& \max_{j}\frac{|\vec{v}^*_j - \vec{v}^{(k)}_j|}{\eta},\\
	\label{eq:UnbalancedLemmaB}
	\left|\log \frac{\vec{b}^*_j}{\vec{b}^{(k)}_j} - \frac{\vec{v}^*_j - \vec{u}^{(k)}_j}{\eta}\right|
	&\leq& \max_{i}\frac{|\vec{u}^*_i- \vec{u}^{(k)}_i|}{\eta}.
\end{eqnarray}
\end{Lem}
{The proof is omitted. Refer to that of {\cite[{\bf Lemma 2}]{UOTSinkhorn2020}}.}

\begin{Lem}{}
\label{lem:SupNormOptimalInqquality}
	The maximum norms of the optimal solution $(\vec{u}^*,\vec{v}^*)$ are bounded as
	\begin{equation}
		\max \left \lbrace \|\vec{u}^*\|_\infty, \|\vec{v}^*\|_\infty \right \rbrace \leq 2(\tau+\eta)R.
	\end{equation}
\end{Lem}

\begin{proof}
We use the proof of {\cite[{\bf Lemma 3}]{UOTSinkhorn2020}}. The following inequalities are true.
\begin{eqnarray*}
\|\vec{u}\|_\infty^* \left(\frac{1}{\tau}+\frac{1}{\eta} \right) 
&\leq & \frac{\|\vec{v}^*\|_\infty}{\eta} + R,\\
\|\vec{v}^*\|_\infty 
&\leq & \|\vec{u}^*\|_\infty + \eta R.
\end{eqnarray*}
When $ \|\vec{v}^*\|_\infty {\leq} \|\vec{u}^*\|_\infty $, we have
\begin{equation*}
	\|\vec{u}\|_\infty^* \left(\frac{1}{\tau}+\frac{1}{\eta} \right) \leq  \frac{\|\vec{v}^*\|_\infty}{\eta} + R \leq \frac{\|\vec{u}^*\|_\infty}{\eta} + R \Longleftrightarrow \|\vec{u}^*\|_\infty \leq {\tau} R.
\end{equation*}
When $\|\vec{v}^*\|_\infty {\geq} \|\vec{u}^*\|_\infty$, we have
\begin{equation*}
	\|\vec{u}\|_\infty^* \left(\frac{1}{\tau}+\frac{1}{\eta} \right) \leq  \frac{\|\vec{v}^*\|_\infty}{\eta} + R \leq \frac{\|\vec{u}^*\|_\infty}{\eta} + 2R \Longleftrightarrow \|\vec{u}^*\|_\infty \leq 2 \tau R.
\end{equation*}
Also, we have
\begin{equation*}
	\|\vec{v}^*\|_\infty  \leq  \|\vec{u}^*\|_\infty + \eta R \leq 2\tau R + \eta R \leq (2\tau+\eta)R.
\end{equation*}
Combining these two inequalities yields the desired result.
\end{proof}

We now provide the proof of {\bf Theorem \ref{thm:AlgorithmCnvergenceEnsure}}.
\begin{proof}
First, we address the case in which $k$ is even. (\ref{eq:SinkhornUpdateEven}) is reformulated as
\begin{align*}
	\vec{u}^{(k+1)}_i 
	=\frac{\tau}{\eta+\tau}\left(\vec{u}^{(k)}_i+\eta (\log (\vec{a}_i) -\log (\vec{a}^*_i)+\log(\vec{a}^*_i )- \log (\vec{a}^{(k)}_i))\right).
\end{align*}
From {\bf Lemma \ref{lem:OptimalEqualityDual}}, the following holds.
\begin{equation*}
	\vec{u}^{(k+1)}_i - \vec{u}^*_i = \frac{\tau}{\tau+\eta}(\vec{u}^{(k)}_i - \vec{u}^*_i +\eta(\log(\vec{a}^*_i)- \log (\vec{a}^{(k)}_i))).
\end{equation*}
{\bf Lemma \ref{lem:UnbalancedLemma2}} yields the following inequality
\begin{equation*}
	|\vec{u}^{(k+1)}_i - \vec{u}^*_i | \leq \frac{\tau}{\tau+\eta}\max_{1\leq j\leq n}|\vec{v}^{(k)}_j - \vec{v}^*_j |.
\end{equation*}
Consequently, we obtain as
\[
	\|\vec{u}^{(k+1)} - \vec{u}^* \|_\infty \leq \frac{\tau}{\tau+\eta}\|\vec{v}^{(k)} - \vec{v}^*\|_\infty. 
\]
Similarly, the inequality $\|\vec{v}^{(k)} - \vec{v}^*\|_\infty \leq \|\vec{u}^{(k-1)} - \vec{u}^* \|_\infty$ is derived. Consequently, combining the two inequalities yields
\begin{equation*}
\|\vec{u}^{(k+1)} - \vec{u}^* \|_\infty \leq \frac{\tau}{\tau+\eta}\|\vec{u}^{(k-1)} - \vec{u}^*\|_\infty.
\end{equation*}
Telescoping this inequality, we obtain
\begin{equation*}
	\|\vec{u}^{(k+1)} - \vec{u}^* \|_\infty  \leq  \frac{\tau}{\tau+\eta}\|\vec{u}^{(k-1)} - \vec{u}^*\|_\infty \leq  \left(\frac{\tau}{\tau+\eta}\right)^{ \frac{k}{2} + 1} \|\vec{v}^{(0)} - \vec{v}^*\|_\infty \leq  \left(\frac{\tau}{\tau+\eta}\right)^{ \frac{k}{2} + 1 } \| \vec{v}^*\|_\infty.
\end{equation*}
Addressing $\vec{v}^{(k+1)} = \vec{v}^{(k)}$, one obtains
\begin{eqnarray*}
	\|\vec{v}^{(k+1)} \!\!- \vec{v}^* \|_\infty =   \|\vec{v}^{(k)} - {\vec{v}^*} \|_\infty  \leq \|\vec{u}^{(k-1)}\!\! - \vec{u}^* \|_\infty  \leq \left(\frac{\tau}{\tau+\eta}\right)^{ \frac{k}{2}  } \| \vec{v}^*\|_\infty.
\end{eqnarray*}
Similarly, when $k$ is odd, one obtains $\|\vec{v}^{(k+1)} - \vec{v}^*\|_\infty \leq \|\vec{u}^{(k)} - \vec{u}^* \|_\infty$ and $\|\vec{u}^{(k)} - \vec{u}^* \|_\infty \leq \frac{\tau}{\tau+\eta}\|\vec{v}^{(k-1)} - \vec{v}^*\|_\infty$. Combining and telescoping them also yields the following inequality:
\begin{eqnarray*}
	&&\|\vec{v}^{(k+1)} - \vec{v}^*\|_\infty   \leq  \frac{\tau}{\tau+\eta}\|\vec{v}^{(k-1)} - \vec{v}^*\|_\infty  \leq  \left(\frac{\tau}{\tau+\eta}\right)^{ \frac{k+1}{2}  } \| \vec{v}^*\|_\infty \\
	&&\|\vec{u}^{(k+1)} - \vec{u}^* \|_\infty = \|\vec{u}^{(k)} - \vec{u}^* \|_\infty \leq \frac{\tau}{\tau + \eta}\|\vec{v}^{(k-1)} - \vec{u}^* \|_\infty \left(\frac{\tau}{\tau+\eta}\right)^{ \frac{k+1}{2}  } \| \vec{v}^*\|_\infty 
\end{eqnarray*}
By combining the two inequalities and by applying {\bf Lemma \ref{lem:SupNormOptimalInqquality}} and {$0 < \frac{\tau}{\tau + \eta} < 1$}, we obtain 
\begin{eqnarray*}
	{\max \left \lbrace \| \vec{u}^{(k+1)}-\vec{u}^{*}\|_\infty,\|\vec{v}^{(k+1)}-\vec{v}^{*}\|_\infty \right \rbrace \notag  \leq 2(\tau+\eta)R \left(\frac{\tau}{\tau+\eta}\right)^{ \frac{k}{2}  } R \leq 2\tau R \left(\frac{\tau}{\tau+\eta}\right)^{ \frac{k}{2} - 1 }.}
\end{eqnarray*}
{Therefore, the inequality (\ref{eq:ConvergenceInequality}) holds}. This completes the proof.
\end{proof}

\subsection{Proof of necessary lemmas for {\bf Theorem \ref{thm:AlgorithmTimeComplexity}}}
{We first  redescribe {\bf Lemma \ref{lem:NewStoppingCreterion}} in the main material, and give its proof}.

\noindent {\bf Lemma 4.3.} {\it When the iteration $k$ is even, the $k$-th matrix $\mat{T}^{(k)}$ satisfies the following inequality.}
\begin{equation*}
\left|\!\left|\log \left ( \frac{\mat{T}^{(k)}}{\mat{T}^{*}}\right) \right|\!\right|_\infty \leq \frac{4\tau}{\eta}R \left (\frac{\tau}{\tau + \eta}\right )^{\frac{{k-1}}{2}-1},
\end{equation*}
{\it where $R = \max \lbrace \| \log (\vec{a} )\|_{{\infty}} , \| \log(\vec{b})\|_{{\infty}} \rbrace + \max  \left\{ \log(n),\frac{1}{\eta}\| \mat{C}\|_\infty - \log(n) \right\}.$}

\begin{proof}
Consider the ratio between the $k$-th matrix $\mat{T}^{(k)}_{ij}$ and the optimal solution $\mat{T}^*_{ij}.$
\begin{equation*}
	\frac{\mat{T}^{(k)}_{i,j}}{\mat{T}^*_{i,j}}=\frac{\exp\left(\frac{\vec{u}^{(k)}_i+\vec{v}^{(k)}_j - \mat{C}_{i,j}}{\eta}\right)}{\exp\left(\frac{\vec{u}^{*}_i+\vec{v}^{*}_j - \mat{C}_{i,j}}{\eta}\right)} = \exp\left(\frac{\vec{u}^{(k)}_i-\vec{u}^*_i+\vec{v}^{(k)}_j - \vec{v}^*_j}{\eta}\right).
\end{equation*}
We then take the logarithm of this ratio and consider the absolute value of it. Then, we obtain 
\begin{eqnarray*}
\left|\log \left (\frac{\mat{T}^{(k)}_{i,j}}{\mat{T}^*_{i,j}}\right)\right|
& = &\frac{|\vec{u}^{(k)}_i-\vec{u}^*_i+\vec{v}^{(k)}_j - \vec{v}^*_j|}{\eta}\\
&\leq& \frac{|\vec{u}^{(k)}_i-\vec{u}^*_i|+|\vec{v}^{(k)}_j - \vec{v}^*_j|}{\eta} \\
&\leq&\frac{\|\vec{u}^{(k)}-\vec{u}^*\|_\infty+\|\vec{v}^{(k)} - \vec{v}^*\|_\infty}{\eta}\\
&\leq & 2\frac{\max \left \lbrace \parallel \vec{u}^{(k)}-\vec{u}^{*}\parallel_\infty,\parallel\vec{v}^{(k)}-\vec{v}^{*}\parallel_\infty \right \rbrace}{\eta}\\
&\leq&\frac{4\tau}{\eta}R \left (\frac{\tau}{\tau + \eta}\right )^{ {\frac{k-1}{2}}  - 1} .
\end{eqnarray*}
The second inequality is derived from the definition of the maximum norm, and the last inequality uses {\bf Theorem \ref{thm:AlgorithmCnvergenceEnsure}}. For all index $i,j$, the inequality above holds. 
{Therefore, we obtain the desired result}.
This completes the proof. 
\end{proof}
Next, we restate {\bf Lemma \ref{lem:LogarithmInequality}} in the main material, and give its proof.  

\noindent{\bf Lemma 4.4.} {\it For {$0 < y < x < b$}}, the logarithm function satisfies the following.
\begin{equation}
\label{eq:mean_val_thm}
	\log x - \log y\geq \frac{1}{b}(x - y).
\end{equation}

\begin{proof}
For this interval, the logarithm function is differentiable and, thereby, the mean-value theorem is applicable to it. {Thus, for $ 0 < y < x < b$}, the following inequality holds.
\begin{equation}
\label{eq:mean_val_equality}
	\frac{\log x - \log y}{x-y} = (\log x)^{\prime}|_{x=c}=\frac{1}{c}\geq \frac{1}{b},
\end{equation} 
where $c$ is in $(x,y)$ and the first inequality is derived from the monotonicity of $\frac{1}{x}$ and $c \leq b$. Because $x$ is greater than $y$, $x - y$ is positive. Then, multiplying $x - y$ by both sides, the sign of the inequality (\ref{eq:mean_val_equality}) is not changeable. Therefore, we obtain the desired result. This completes the proof.
\end{proof}

{From this lemma, we obtain related inequalities in terms of the absolute value as well as the maximum norm.}
{\begin{Cor}[]
\label{AppenCor:maximum_norm_key_inequality}

{For $x , y \in (0,b)$, the logarithm function satisfies the following}
\begin{equation}
	\label{eq:mean_val_thm_abs}
	|\log x - \log y| \geq \frac{1}{b} |x - y|.
\end{equation} 
Furthermore, for $\vec{x},\vec{y} \in \mathbb{R}^{{n}}$ that satisfy $\vec{x}_i,\vec{y}_i \in (0,b)$ for all $i$, the following inequality holds:
\begin{equation}
	\label{eq:mean_val_thm_maxnorm}
	\|\log (\vec{x}) - \log (\vec{y}) \|_\infty  \geq \frac{1}{b} \|\vec{x} - \vec{y}\|_\infty.
\end{equation}
\end{Cor}}

{\begin{proof}
Because a logarithm function is monotone, $\log x < \log y $ holds when $x < y$, and vice versa. Then, {the sign of  both sides of the inequality (\ref{eq:mean_val_equality}) is positive}. Thereby, (\ref{eq:mean_val_thm}) holds for the absolute value. Furthermore, if $\vec{x},\vec{y} \in \mathbb{R}^{{n}}$ satisfy $\vec{x}_i,\vec{y}_i \in (0,b)$ for all $i$, their elements satisfy the inequality (\ref{eq:mean_val_thm_abs}) in {\bf Corollary \ref{AppenCor:maximum_norm_key_inequality}} . This means that (\ref{eq:mean_val_thm}) holds when evaluating in the maximum norm. 
\end{proof}}

{We finally redescribe {\bf Lemma \ref{lem:KeyInequality}} in the main material, and give its proof}.

\noindent {{\bf Lemma 4.5.} {\it The iteration $k$ is even after the odd update. The optimal solution $\mat{T}^*$ and the $k$-th iterate $\mat{T}^{(k)}$ generated by the SR--Sinkhorn algorithm satisfy the following inequality.}
\begin{equation*}
	\|\log \mat{T}^{(k)} - \log \mat{T}^*\|_\infty \geq \frac{1}{\beta} \|\mat{T}^{(k)} - \mat{T}^* \|_\infty.
\end{equation*}}

\begin{proof}
Because the matrices $\mat{T}^{(k)}$ and $\mat{T}^*$ satisfy the primal marginal constraint $\mat{T}^T\vec{1}_n = \vec{b}$, {any elements of $\mat{T}^{(k)}$ and $\mat{T}^*$} satisfy $0 \leq \mat{T}^{(k)}_{i,j}, \mat{T}^*_{i,j} \leq \beta$ for all $i,j$. Therefore, (\ref{eq:mean_val_thm_maxnorm}) in {\bf Corollary \ref{AppenCor:maximum_norm_key_inequality}} is applicable. Then, we obtain
\begin{equation*}
	\|\log \mat{T}^{(k)} - \log \mat{T}^*\|_\infty \geq \frac{1}{\beta} \|\mat{T}^{(k)} - \mat{T}^* \|_\infty.
\end{equation*}
This completes the proof.
\end{proof}

\subsection{Full proof of Theorem \ref{thm:AlgorithmTimeComplexity}}
{We first restate {\bf Theorem \ref{thm:AlgorithmTimeComplexity}} in the main material, and give its proof}. 

\noindent
{\bf Theorem 4.1} (convergence to $\epsilon$-approximation {w.r.t.} functional value {gap}). {\it Letting $c_1$ and $c_2$ respectively represent $(\frac{2n(\tau+\eta)R}{\tau} + 1) \beta$} and $2\beta \log n$, then one can consider the case in which $k$ is even after the odd update. If $\|\log \left ( \frac{{\bf T}^{(k)}}{{\bf T}^{*}}\right) \|_\infty \leq \epsilon^{\prime}$, $\mat{T}^{(k)}$ generated by the SR--Sinkhorn algorithm satisfies
\begin{equation*}
	f(\mat{T}^{(k)})-f(\hat{\mat{T}}) \leq (\beta \|\mat{C}\|_1 + \tau c_1)\epsilon^{\prime}+ \eta c_2.
\end{equation*}
Furthermore, defining $\epsilon^{\prime} = \frac{\epsilon_f}{2(\beta\|{\bf {C}}\|_1 +\tau c_1)}$ and $\eta = \frac{\epsilon_f}{2c_2}$ for an approximation constant $\epsilon_f$, $f(\mat{T}^{(k)})-f(\hat{\mat{T}}) \leq \epsilon_f$ holds.

We now present the full proof of {\bf Theorem \ref{thm:AlgorithmTimeComplexity}.}

\begin{proof}
We assume that the number iteration $k$ is even. Then, we have
\begin{align*}
f(\mat{T}^{(k)})-f(\hat{\mat{T}})&= g(\mat{T}^{(k)})-g(\hat{\mat{T}})+\eta(\mathrm{H}(\mat{T}^{(k)})-\mathrm{H}(\hat{\mat{T}}))\\
&= g(\mat{T}^{(k)})-g(\hat{\mat{T}})+g(\mat{T}^*)-g(\mat{T}^*)+\eta(\mathrm{H}(\mat{T}^{(k)})-\mathrm{H}(\hat{\mat{T}}))\\
&\leq g(\mat{T}^{(k)})-g(\mat{T}^*)+\eta(\mathrm{H}(\mat{T}^{(k)})-\mathrm{H}(\hat{\mat{T}}))\\
&\leq \langle  \mat{C},\mat{T}^{(k)} - \mat{T}^* \rangle + \tau (\mathrm{KL}(\mat{T}^{(k)}\vec{1}_n,\vec{a}) - \mathrm{KL}(\mat{T}^*\vec{1}_n,\vec{a}))+\eta(\mathrm{H}(\mat{T}^{*})-\mathrm{H}(\hat{\mat{T}})),
\end{align*}
where the first inequality uses $ g(\mat{T}^*)\leq g(\hat{\mat{T}})$ because $\mat{T}^*$ is the optimal solution of $g(\mat{T})$. The upper bounds of these three terms are separately considered as shown below.

\noindent{\bf Upper bound of $\langle \mat{C}, \mat{T}^{(k)} - \mat{T}^* \rangle$: }The term is bounded by the Holder's inequality as
\begin{eqnarray*}
	\langle  \mat{C} , \mat{T}^{(k)} - \mat{T}^* \rangle \leq  \|\mat{C}\|_1\|\mat{T}^{(k)} - \mat{T}^* \|_\infty \leq \beta \|\mat{C}\|_1\| \log\mat{T}^{(k)} - \log\mat{T}^*\|_\infty,
\end{eqnarray*}
where the second inequality uses {\bf Lemma \ref{lem:KeyInequality}}. Here, we consider the stopping criterion in {\bf Lemma \ref{lem:NewStoppingCreterion}}. Consequently, the term is bounded by
\begin{equation}
\label{eq:bound_1}
\langle \mat{C} ,\mat{T}^{(k)} - \mat{T}^*\rangle \leq \beta \|\mat{C}\|_1\epsilon^{\prime}.
\end{equation}

\noindent{\bf Upper bound of $\mathrm{H}(\mat{T}^*) - \mathrm{H}(\hat{\mat{T}})$: }We utilize the inequality of the paper {\cite[Eq.(15)]{UOTSinkhorn2020}}. The first term $\mathrm{H}(\mat{T}^*)$ is bounded by
\begin{equation*}
	\mathrm{H}(\mat{T}^*)\leq 2 t^* \log n  + t^* - t^*\log t^*=2\beta\log n + \beta - \beta \log \beta,
\end{equation*}
where {$t^*=\sum_{i,j} \mat{T}^*_{i,j}=\beta$, and} where the optimal solution $\mat{T}^*$ satisfies the constraint $\mat{T} \geq \mat{0}, \mat{T}^T\vec{1}_n = \vec{b}$. The second term $\mathrm{H}(\hat{\mat{T}})$ is bounded in another way by
\begin{equation*}
	-\mathrm{H}(\hat{\mat{T}}) \leq   \hat{t}\log \hat{t} - \hat{t}=\beta \log\beta - \beta,
\end{equation*}
where {$\hat{t}=\sum_{i,j} \hat{\mat{T}}_{i,j}=\beta$.}
From both terms, we obtain 
\begin{equation*}
	\mathrm{H}(\mat{T}^{*})-\mathrm{H}(\hat{\mat{T}}) \leq  2\beta\log n =: c_2.
\end{equation*}

\noindent{\bf Upper bound of $\mathrm{KL}(\mat{T}^{(k)}\vec{1}_n,\vec{a}) - \mathrm{KL}(\mat{T}^*\vec{1}_n,\vec{a})$: }We denote {$\mat{T}^{(k)}\vec{1}_n$} and $\mat{T}^*\vec{1}_n$ as $\vec{a}^{(k)}$ and $\vec{a}^*$, respectively. We rearrange this term as
\begin{eqnarray*}
	&&\mathrm{KL}(\mat{T}^{(k)}\vec{1}_n,\vec{a}) - \mathrm{KL}(\mat{T}^{*}\vec{1}_n,\vec{a}) \\
	& = &\sum_i \left ( \vec{a}^{(k)}_i \log \frac{\vec{a}^{(k)}_i}{\vec{a}_i}-\vec{a}^{(k)}_i +\vec{a}_i \right) - \sum_i \left(\vec{a}^{*}_i \log \frac{\vec{a}^{*}_i}{\vec{a}_i}-\vec{a}^{*}_i + \vec{a}_i \right) \\
	&=& \sum_i \left ( \vec{a}^{(k)}_i \log \frac{\vec{a}^{(k)}_i}{\vec{a}_i} \right) - \sum_i \left(\vec{a}^{*}_i \log \frac{\vec{a}^{*}_i}{\vec{a}_i} \right)  + \sum_i \vec{a}^{(k)}_i - \sum_i \vec{a}^*_i\\
	&=& \sum_i \left ( \vec{a}^{(k)}_i \log \frac{\vec{a}^{(k)}_i}{\vec{a}_i} \right) - \sum_i \left(\vec{a}^{*}_i \log \frac{\vec{a}^{*}_i}{\vec{a}_i} \right) \\
	&=& \sum_i \left ( (\vec{a}^{*}_i - \vec{a}^{(k)}_i)\log \vec{a}_i \right) + \sum_i \left ( \vec{a}^{(k)}_i \log \vec{a}^{(k)}_i- \vec{a}^{*}_i \log \vec{a}^{*}_i\right ) \\
	&=& \sum_i \left ( (\vec{a}^{*}_i - \vec{a}^{(k)}_i)\log \vec{a}_i \right) + \sum_i \left ( \vec{a}^{(k)}_i (\log \vec{a}^{(k)}_i-\log \vec{a}^{*}_i) \right ) + \sum_i \left ( (\vec{a}^{(k)}_i - \vec{a}^{*}_i)\log \vec{a}^{*}_i \right)\\
	&=&\sum_i \left ( (\vec{a}^{(k)}_i - \vec{a}^{*}_i)(\log \vec{a}^{*}_i - \log \vec{a}_i )\right) + \sum_i \left ( \vec{a}^{(k)}_i (\log \vec{a}^{(k)}_i-\log \vec{a}^{*}_i) \right ),
\end{eqnarray*}
where the second equality uses $\sum_i \vec{a}^{(k)}_i = \sum_i \vec{a}^{*}_i = \beta$. Following this, we bound this term as
\begin{eqnarray*}
	&&\sum_i \left ( (\vec{a}^{(k)}_i - \vec{a}^{*}_i)(\log \vec{a}^{*}_i - \log \vec{a}_i )\right) + \sum_i \left ( \vec{a}^{(k)}_i (\log \vec{a}^{(k)}_i-\log \vec{a}^{*}_i) \right ) \\
	&\leq&\sum_i |  (\vec{a}^{(k)}_i - \vec{a}^{*}_i)(\log \vec{a}^{*}_i - \log \vec{a}_i )| + \sum_i |\vec{a}^{(k)}_i (\log \vec{a}^{(k)}_i-\log \vec{a}^{*}_i)| \\
	&\leq&\sum_i |  \vec{a}^{(k)}_i - \vec{a}^{*}_i|\cdot |\log \vec{a}^{*}_i - \log \vec{a}_i | + \sum_i |\vec{a}^{(k)}_i |\cdot|\log \vec{a}^{(k)}_i-\log \vec{a}^{*}_i| \\
	&\leq& {\sum_i  \beta |  \log \vec{a}^{(k)}_i - \log \vec{a}^{*}_i|\cdot |\log \vec{a}^{*}_i - \log \vec{a}_i |+ \sum_i |\vec{a}^{(k)}_i |\cdot|\log \vec{a}^{(k)}_i-\log \vec{a}^{*}_i| }\\
	&\leq& {\sum_i  \beta  \frac{4\tau}{\eta}R\left (\frac{\tau}{\tau+\eta}\right )^{ \frac{{k-1}}{2}  -1} \cdot |\log \vec{a}^{*}_i - \log \vec{a}_i |+ \sum_i |\vec{a}^{(k)}_i |\cdot \frac{4\tau}{\eta}R\left (\frac{\tau}{\tau+\eta}\right )^{ \frac{{k-1}}{2}  -1} }\\
	&\leq& {\frac{4\tau}{\eta}R\left (\frac{\tau}{\tau+\eta}\right )^{ \frac{{k-1}}{2}  -1} \left ( \sum_i \beta \frac{\|\vec{u}^*\|_\infty}{\tau} + \sum_i |\vec{a}^{(k)}_i| \right) }\\
	&\leq& \left(\frac{2n(\tau+\eta)R}{\tau} + 1 \right) \beta \epsilon^\prime := c_1 \epsilon^\prime,
\end{eqnarray*}
where the first inequality is derived from $x \leq |x|$ and the third inequality uses (\ref{eq:mean_val_thm_abs}) in {\bf Corollary \ref{AppenCor:maximum_norm_key_inequality}}. 
The fourth and fifth inequalities use {\bf Lemma \ref{lem:UnbalancedLemma2}} and {\bf Lemmas \ref{lem:OptimalEqualityDual}}, respectively. The sixth inequality is transformed by {\bf Lemma \ref{lem:SupNormOptimalInqquality}}. The last equality can be rearranged as the same as the case of the upper bound of $\langle \mat{T}^{(k)} - \mat{T}^*, \mat{C} \rangle$.

From {these three upper bounds}, we can bound the KL and entropy terms as $\epsilon^{\prime}c_1$ and $c_2$ respectively. Finally, putting all of them together yields
\begin{eqnarray*}
	f(\mat{T}^{(k)})-f(\hat{\mat{T}}) &\leq& \beta \|\mat{C}\|_1\epsilon^{\prime} + \tau c_1 \epsilon^{\prime}+ \eta c_2.
\end{eqnarray*}
Setting $\eta = \frac{\epsilon_f}{2c_2}, \epsilon^{\prime} \!=\! \frac{\epsilon_f}{2(\beta\|{\bf {C}}\|_1+\tau c_1)}$, $f(\mat{T}^{(k)})-\!f(\hat{\mat{T}}) $ is bounded by the approximation constant $\epsilon_f$. This completes the proof.
\end{proof}

\subsection{Convergence rate and total complexity}
This section gives  the total complexity, considering a specific stopping criterion to satisfy an $\epsilon$-approximation according to {\bf Lemma \ref{lem:NewStoppingCreterion}}.
We first restate {\bf Corollary \ref{cor:ComplexityAndOrder}} in the main material.

\noindent{\bf Corollary 4.2} (stopping iteration bound and total complexity).
{\it Letting $c_1$ and $c_2$ respectively denote $(\frac{2n(\tau+\eta)R}{\tau} + 1) \beta$ and $2\beta \log n$, then {\bf Theorem \ref{thm:AlgorithmTimeComplexity}} holds. The stopping iteration bound is given as}
\begin{equation*}
	 k \geq 2  \Bigl(1+\frac{2c_2\tau}{\epsilon_f}\Bigr) \Bigl(\log 16\tau R  + \log c_2(\beta\|\mat{C}\|_1+ \tau c_1)  + 2\log \frac{1}{\epsilon_f} \Bigr) + 3.
\end{equation*}
{\it Furthermore, assuming $R = \mathcal{O}(\frac{1}{\eta}\|\mat{C}\|_\infty)$, the total complexity of the SR--Sinkhorn algorithm  is }
\begin{equation*}
	\mathcal{O}\left ( \frac{\tau n^2}{\epsilon_f} \log n  \left (  \log n + \log \tau + \log (\log n) + \log \|\mat{C}\|_\infty + \log (n^2  + \frac{n \tau \log n}{\epsilon_f}) + \log \frac{1}{\epsilon_f} \right ) \right).
\end{equation*}

We now gives the proof of {\bf Corollary \ref{cor:ComplexityAndOrder}}.
\begin{proof}
From {{\bf Lemma \ref{lem:NewStoppingCreterion}}}, we have
\begin{equation*}
	\frac{4\tau }{\eta}R \left (\frac{\tau}{\tau + \eta} \right)^{\frac{{k-1}}{2} -1}  \leq \epsilon^{\prime}.
\end{equation*}
Taking logarithm of both sides of this gives
\begin{equation*}
	\label{AppenEq:InequalityRate}
	\left(\frac{{k-1}}{2} - 1 \right)\log \left(1 - \frac{\tau}{\tau+\eta}\right) 
	\leq 
	- {\log \left ( \frac{4\tau R}{\epsilon^\prime \eta}\right) }
\end{equation*}
Herein, because the inequality $\log (1 - x) \geq -x$ for $x \in (0,1)$, the inequality is 
\begin{equation*}
	k \geq 3 + 2\left(1+\frac{\tau}{\eta}\right) \left(\log (4\tau R) + \log \left( \frac{1}{\epsilon^\prime} \right) +  \log \left( \frac{1}{\eta} \right) \right).
\end{equation*}
Substituting $\eta = \frac{\epsilon_f}{2c_2}$ and $\epsilon^{\prime} = \frac{\epsilon_f}{2(\beta\|{\bf C}\| + \tau c_1)}${, we obtain}
\begin{equation}
	k \geq 3 + 2\left(1+\frac{2c_2\tau}{\epsilon_f}\right) \left(\log (16\tau R) + \log \left( c_2 \right) +  \log \left(\beta\|\mat{C}\|_1 + c_1\tau \right) + 2\log \frac{1}{\epsilon_f}\right).
\end{equation}

Next, we consider the complexity of the iteration. Because the $\eta$ is $\frac{\epsilon_f}{2\beta \log n}$ and the order of the term $R$ is $\mathcal{O}(\frac{1}{\eta}\|\mat{C}\|_\infty) = \mathcal{O}(\frac{\log n}{\epsilon_f}\|\mat{C}\|_\infty)$, the order of the constant $\beta \|\mat{C}\|_1 + \tau c_1$ is transform into
\begin{eqnarray*}
	\beta \|\mat{C}\|_1 + \tau c_1& = &\beta \|\mat{C}\|_1  + \tau \beta (\frac{2n(\tau+\eta)R}{\tau} + 1)  \\
	&=&\mathcal{O} \left (  \beta n^2 \|\mat{C}\|_\infty + \beta 2n \frac{\tau+\eta}{\eta}\|\mat{C}\|_\infty \right) \\
	&=& \mathcal{O} \left (  \beta \|\mat{C}\|_\infty ( n^2  +  2n (1+\frac{\tau \beta \log n}{\epsilon_f} ) )\right).
\end{eqnarray*}

Combining the matrix operation $\mathcal{O}(n^2)$, the time complexity is {given as}
\begin{equation}
	\mathcal{O}\left ( \frac{\tau n^2}{\epsilon_f} \log n  \left (  \log n + \log \tau + \log (\log n) + \log \|\mat{C}\|_\infty + \log (n^2  + \frac{n \tau \log n}{\epsilon_f}) + \log \frac{1}{\epsilon_f} \right ) \right).
\end{equation}
This completes the proof. 
\end{proof}

\section{Theoretical results about the marginal constraint {gap}}
This section presents theoretical results about the marginal constraint gap. We first provide the proof of {{\bf Theorem \ref{thm:MarginalConstraintError}}, i.e.,} the convergence rates in the marginal constraint {gap}s of not only the non-relaxed constraint of the vector $\vec{b}$ but also the relaxed constraint of $\vec{a}$. After that, we bound the second term of the inequality (\ref{eq:marginal_bound_a}) by given parameters on the conditions where two distributions $\vec{a}$ and $\vec{b}$ are the probability simplex $\Delta^n$. We then finally provide the proof of {{\bf Theorem \ref{thm:AlgorithmTimeMarginalComplexity}},} the convergence rate to the $\epsilon$-approximation in terms of the marginal constraint gap.

\subsection{Proof  of {\bf Theorem \ref{thm:MarginalConstraintError}}}

{We first redescribe {\bf Theorem \ref{thm:MarginalConstraintError}} in the main material}. 

\noindent{\bf Theorem 4.7} (convergence rates of marginal constraint {gap})
{\it One can consider the case for even $k$ after the odd update. Then, the marginal constraint {gap} of the vector $\vec{a}$ satisfies}
\begin{eqnarray}
	\label{eq:marginal_bound_a}
	\|\vec{a}^{(k)} - \vec{a}\|_\infty \leq \gamma \left ( \frac{4\tau}{\eta}R\left (\frac{\tau}{\tau+\eta}\right )^{ \frac{{k-1}}{2}  -1}  + \frac{\|\vec{u}^*\|_\infty}{\tau}\right),
\end{eqnarray}
{\it where $\gamma = \max \lbrace \alpha,\beta \rbrace$ and $\vec{u}^*$ is the optimal solution of the dual form of (\ref{eq:EntropySemiRelaxed}). Furthermore, assuming that $k$ is odd after the even update, the logarithm marginal gap of $\vec{b}$ is bounded by}
\begin{equation}
\label{eq:marginal_bound_b}
	\| \log(\vec{b}^{(k)}) - \log(\vec{b}) \|_\infty \leq \frac{4\tau}{\eta}R\left (\frac{\tau}{\tau+\eta}\right )^{ \frac{{k-1}}{2}  -1} .
\end{equation}

We now prove the convergence rates in the marginal constraint gap of {\bf Theorem \ref{thm:MarginalConstraintError}}.
\begin{proof}
Following the proof of {\bf Lemma \ref{lem:UnbalancedLemma2}}, we can prove 
\begin{equation*}
	\left|\log \frac{\vec{a}^{(k)}_i}{\vec{a}^{{*}}_i} - \frac{\vec{u}^{(k)}_i - \vec{u}^{*}_i}{\eta}\right| \leq \max_{j}\frac{|\vec{v}^{(k)}_j- \vec{v}^{*}_j|}{\eta},
\end{equation*}
where $k$ is even. We have
\begin{eqnarray}
\label{eq:log_a_a}
	\left| \log \frac{\vec{a}^{(k)}_j}{\vec{a}^*_j} \right | = 
	&\leq &
	\max_i \frac{|\vec{u}^{(k)}_i - \vec{u}^{*}_i|}{\eta} +  \max_{j}\frac{|\vec{v}^{(k)}_j- \vec{v}^{*}_j|}{\eta} \notag\\
	&\leq &
	\frac{2}{\eta} \max \lbrace \|\vec{u}^{(k)} - \vec{u}^{*}\|_\infty,\|\vec{u}^{(k)} - \vec{u}^{*}\|_\infty \rbrace \notag\\
	&\leq & \frac{4\tau}{\eta} R\left (\frac{\tau}{\tau + \eta}\right )^{{ \frac{k-1}{2} }-1},
\end{eqnarray}
where the last equality uses {\bf Theorem \ref{thm:AlgorithmCnvergenceEnsure}}. {Note} that this also holds for the maximum norm. Also, since $k$ is even, the sum of the elements of $\vec{a}^*,\vec{a}^{(k)}$ is equal to $\beta$. Considering the sum of elements of $\vec{a}$ is equal to $\alpha$, we define the constant $\gamma =  \max \lbrace \alpha,\beta \rbrace$. {By doing so, we obtain} the constraints as $\vec{a}^*_i,\vec{a}_i,\vec{a}^{(k)}_i \leq \gamma$. {Consequently, $\|\vec{a}^{(k)} - \vec{a}\|_\infty$ is upper-bounded as}
\begin{eqnarray*}
	\|\vec{a}^{(k)} - \vec{a}\|_\infty
	& \leq & \gamma \|\log (\vec{a}^{(k)}) - \log (\vec{a}) \|_\infty\\
	& = & \gamma \|\log (\vec{a}^{(k)}) - \log (\vec{a}^{*})+ \log (\vec{a}^{*}) - \log (\vec{a}) \|_\infty \\
	&\leq& \gamma \left ( \|\log (\vec{a}^{(k)}) - \log (\vec{a}^{*})\|_\infty+ \| \log (\vec{a}^{*}) - \log (\vec{a}) \|_\infty \right) \\
	& \leq & \gamma \left ( \frac{4\tau}{\eta}R \left ( \frac{\tau}{\tau+\eta} \right)^{{\frac{k-1}{2}}-1} + \frac{\|\vec{u}^*\|_\infty}{\tau} \right),
\end{eqnarray*}
where the first inequality uses (\ref{eq:mean_val_thm_maxnorm}) in {\bf Corollary  \ref{AppenCor:maximum_norm_key_inequality}}, and the second inequality is derived from the triangle inequality. {Also, the last inequality uses {\bf Lemmas \ref{lem:OptimalEqualityDual}} together with the inequality (\ref{eq:log_a_a}).} This completes the first statement.

As for the second part, repeating the same discussion of (\ref{eq:log_a_a}) for $\vec{b}^{(k)}$ and considering $\vec{b} = \vec{b}^*$, we obtain the desired result. This completes the proof. 
\end{proof}

\subsection{Bounding the maximum norm of the dual form}
We now provide the upper bound of the maximum norm of the dual optimal {solution $\vec{u}^*$. For this purpose, we first give the following lemma.}
\begin{Lem}
\label{lem:element_lemma}
Assume that $\vec{a},\vec{b} \in \Delta^n$. For the {dual} optimal solution $\vec{u}^*$, there exist two different indices $l_1$ and $l_2$ {that} satisfy $\vec{u}^*_{l_1} > 0 \land \vec{u}^*_{l_2} < 0$. Otherwise, $\vec{u}^* = \vec{0}$.
\end{Lem}
\begin{proof}
	We refer to the proof of \cite[{\bf Lemma E.2}]{extrapolation2022Quang}. We will show {that a index $l_1$ exists such that $\vec{u}^*_{l_1} \geq 0$}. Assume for the sake of contradiction that $ \vec{u}^*_{i} < 0$ for all $i$ $\in [n]$. Then from {\bf Lemma \ref{lem:OptimalEqualityDual}} and $\vec{a}_i, \vec{a}^*_i \leq 1$, we have 
	\begin{eqnarray*}
		  0 > \frac{\vec{u}^*_i}{\tau} = \log \vec{a}_i - \log \vec{a}^*_i 
		  &\Longleftrightarrow& \log \vec{a}_i < \log \vec{a}^*_i \Longleftrightarrow  \vec{a}_i <  \vec{a}^*_{i}. 
	\end{eqnarray*}
Summing both sides of the inequality from $1$ to $n$ yields
\begin{equation*}
	  1 = \alpha = \sum_i \vec{a}_i   < \sum_{i} \vec{a}^*_i = \beta = 1.
\end{equation*}
This contradicts $\alpha = \beta = 1$. Therefore, there exists an index $l_1$ such that $\vec{u}^*_{l_1} \geq 0$. Similarly, assuming for the sake of contradiction that $\exists_{i} \vec{u}^*_{i} > 0$, we can show $\exists_{l_2} \vec{u}^*_{l_2} \leq 0$.
	
Next, we take two subsets $S_1,S_2 \subset [n]$ {under the assumption that $S_1 \cup S_2 = [n]$ and $S_1 \cap S_2 = \phi$ hold. In addition, we assume} for the sake of contradiction that the set $S_1$ satisfies $\vec{u}^*_l < 0$ for all $l \in S_1$ and the set $S_2$ satisfies $\vec{u}^*_l = 0$ for all $l \in S_2$. Then, as for the set $S_1$, we have
\begin{equation*}
\sum_{l \in S_1} \vec{a}_l < \sum_{l \in S_1} \vec{a}^*_{l}.
\end{equation*}
Similarly, we also have for the set $S_2$
\begin{equation*}
\sum_{l \in S_2} \vec{a}_l = \sum_{l \in S_2} \vec{a}^*_{l}.
\end{equation*}
These two conditions leads to $1 = \alpha = \sum_i \vec{a}_i < \sum_{i} \vec{a}^*_{i} = \beta = 1$ and this contradicts $\alpha = \beta = 1$. Similarly, assuming for the sake of contradiction that the set $S_1$ satisfies $\vec{u}^*_l > 0$ {for all $l \in S_1$} and the set $S_2$ satisfies $ \vec{u}^*_l = 0$ for all $l \in S_2$, {$1 = \sum_i \vec{a}_i > \sum_{i} \vec{a}^*_{i} = 1$ holds, but this also contradicts $\alpha = \beta = 1$.} Therefore, {there exisits $\vec{u}^*$ {such that} $\vec{u}^{*}_{l_1} > 0$ and $\vec{u}^{*}_{l_2} < 0$ for two different indices or $\vec{u}^* = \vec{0}$.} This completes the proof.
\end{proof}

\begin{Lem}
\label{lem:new_supreme_norm}
When $\vec{a}, \vec{b} \in \Delta^n$, the maximum norm $\|\vec{u}^*\|_\infty$ can be bounded as
	\begin{equation*}
		\|\vec{u}^*\|_\infty \leq \frac{\tau}{\tau + \eta} \left ( \|\mat{C}\|_\infty +\eta(\log \vec{a}_{\rm max} - \log \vec{a}_{\rm min}) \right).
	\end{equation*}
\end{Lem}

We then prove {\bf Lemma \ref{lem:new_supreme_norm}} with the help of {\bf Lemma \ref{lem:element_lemma}}.
\begin{proof}
We can bound $\vec{a}^{(k)}_i$ as
\begin{equation*}
	\vec{a}^*_i = \sum_j \exp \left ( \frac{\vec{u}^*_i + \vec{v}^*_j - \mat{C}_{i,j}}{\eta} \right  )  \geq \sum_j \exp \left(\frac{\vec{u}^*_i + \vec{v}^*_j - \|\mat{C}\|_\infty}{\eta}\right) .
\end{equation*}
Taking logarithm of the both hand of the inequality, we have for {all $i \in [n]$}
\begin{eqnarray}
	\nonumber &&\log \vec{a}^*_i \geq \frac{\vec{u}^*_i}{\eta} - \frac{\|\mat{C}\|_\infty}{\eta} + \log \left(\sum_j \exp\frac{\vec{v}^*_j }{\eta} \right).
\end{eqnarray}	
{Substituting the result of {\bf Lemma \ref{lem:OptimalEqualityDual}} into the left-hand side of this inequality yields}
\begin{eqnarray}	
	\nonumber &&\log \vec{a}_i  - \frac{\vec{u}^*_i}{\tau} \geq \frac{\vec{u}^*_i}{\eta} - \frac{\|\mat{C}\|_\infty}{\eta} + \log \left(\sum_j \exp\frac{\vec{v}^*_j }{\eta}\right).
\end{eqnarray}
{Reformulating this gives}
\begin{eqnarray}	
	\label{eq:Upper}  
	\log \vec{a}_i + \frac{\|\mat{C}\|_\infty}{\eta} - \log \left( \sum_j \exp\frac{\vec{v}^*_j }{\eta} \right) 
	&\geq& \left (\frac{1}{\tau}+ \frac{1}{\eta} \right )\vec{u}^*_i,
\end{eqnarray}
{Considering the case} of $i = l_1$, we obtain
\begin{equation}
	\label{eq:Upper2}  
	\log \vec{a}_{l_1}  + \frac{\|\mat{C}\|_\infty}{\eta} - \log \left (\sum_j \exp\frac{\vec{v}^*_j }{\eta} \right ) \geq  \left (\frac{1}{\tau}+ \frac{1}{\eta} \right )\vec{u}^*_{l_1} \geq 0.
\end{equation} 
In addition, the upper bound of $\vec{a}_i$ can be bounded as
\begin{equation*}
	\vec{a}^*_i = \sum_j \exp\left ( \frac{\vec{u}^*_i + \vec{v}^*_j - \mat{C}_{i,j}}{\eta}\right)   \leq \sum_j \exp \left ( \frac{\vec{u}^*_i + \vec{v}^*_j }{\eta}\right).
\end{equation*}
Similarly, taking logarithm of the both sides of the inequality and using {\bf Lemma \ref{lem:OptimalEqualityDual}}, we have for all $i \in [n]$
\begin{equation}
	\label{eq:Lower}  
	 \log \vec{a}_i   - \log \left (\sum_j \exp\frac{\vec{v}^*_j }{\eta}\right ) \leq  \left (\frac{1}{\tau}+\frac{1}{\eta} \right )\vec{u}^*_i .
\end{equation}
Considering the case of $i = l_2$ also gives
\begin{equation}
	\label{eq:Lower2}  
	\log \vec{a}_{l_2}  -  \log \left (\sum_j \exp\frac{\vec{v}^*_j }{\eta} \right )  \leq \left (\frac{1}{\tau}+ \frac{1}{\eta} \right )\vec{u}^*_{l_2}  \leq 0.
\end{equation}
Therefore, the following inequality holds.
\begin{equation*}
\label{eq:ImportIne}
	\log \vec{a}_{l_2}   \leq \log \left (\sum_j \exp\frac{\vec{v}^*_j }{\eta} \right ) \leq \log \vec{a}_{l_1}  + \frac{\|\mat{C}\|_\infty}{\eta} .
\end{equation*}
{We now derive the upper and lower bounds of $ (\frac{1}{\eta} + \frac{1}{\tau} ) \vec{u}^*_i$. As for the upper bound, putting (\ref{eq:Lower2}) into (\ref{eq:Upper}) yields}
\begin{eqnarray*}
	\left (\frac{1}{\eta} + \frac{1}{\tau} \right ) \vec{u}^*_i &\leq& \log \vec{a}_i  +  \frac{\|\mat{C}\|_\infty}{\eta} - \log \left (\sum_j \exp\frac{\vec{v}^*_j }{\eta} \right ) \\
	&\leq& \log \vec{a}_i  +  \frac{\|\mat{C}\|_\infty}{\eta} - \log \vec{a}_{l_2} \leq \frac{\|\mat{C}\|_\infty}{\eta} + \log \vec{a}_{\rm max} - \log \vec{a}_{\rm min}. 
\end{eqnarray*}
{On the other hand, the lower bound is derived by substituting (\ref{eq:Upper2}) in (\ref{eq:Lower}) as}
\begin{eqnarray*}
	\left (\frac{1}{\eta} + \frac{1}{\tau} \right ) \vec{u}^*_i &\geq& \log \vec{a}_{i}  - \log \left (\sum_j \exp\frac{\vec{v}^*_j }{\eta} \right ) \\
	&\geq& \log \vec{a}_i  - \frac{\|\mat{C}\|_\infty}{\eta} - \log \vec{a}_{l_1} \geq - \frac{\|\mat{C}\|_\infty}{\eta} + \log \vec{a}_{\rm min} - \log \vec{a}_{\rm max}. 
\end{eqnarray*}
From two inequalities, we obtain 
\begin{equation*}
	- \frac{\|\mat{C}\|_\infty}{\eta} + \log \vec{a}_{\rm min} - \log \vec{a}_{\rm max} \leq \left (\frac{1}{\eta} + \frac{1}{\tau} \right ) \vec{u}^*_i \leq \frac{\|\mat{C}\|_\infty}{\eta} + \log \vec{a}_{\rm max} - \log \vec{a}_{\rm min}.
\end{equation*}
{Consequently, the upper bound of the maximum norm of $\vec{u}^*$ is provided as}
\begin{equation*}
\label{Appeneq:New_maximum_norm_bound}
	\| \vec{u}^* \|_\infty \leq \frac{\tau}{\tau + \eta} \left ({\|\mat{C}\|_\infty} + \eta  ( \log \vec{a}_{\rm max} - \log \vec{a}_{\rm min})\right ) .
\end{equation*}
This completes the proof.
\end{proof}

\subsection{Proof of {\bf Theorem \ref{thm:AlgorithmTimeMarginalComplexity}}}

{\bf Theorem 4.9} (convergence to $\epsilon$-approximation of marginal constraint {gap}.)
{\it Assume $\vec{a},\vec{b} \in \Delta^n$. Consider the case in which $k$ is even after the odd update.
Then, the marginal constraint {gap} of the vector $\vec{a}$ can be newly bounded as}
\begin{equation*}
	\|\vec{a}^{(k)} - \vec{a}\|_\infty \leq \frac{4\tau}{\eta}R\left (\frac{\tau}{\tau + \eta}\right )^{ \frac{{k-1}}{2}  -1}  + {\frac{U}{\tau+\eta}},
\end{equation*}
{\it where $\vec{a}_{\rm max}$ and $\vec{a}_{\rm min}$ denote the maximum and minimum element of $\vec{a}$, respectively. $U$ represents $U =  \|\mat{C}\|_\infty+\eta \log \left( \vec{a}_{\rm max}/\vec{a}_{\rm min}\right)$.}
{\it Also if $\|\log \left ( \frac{{\bf T}^{(k)}}{{\bf T}^{*}}\right) \|_\infty \leq \epsilon^{\prime}$, the bound inequality of the marginal constraint gap is given as}
\begin{equation*}
	\|\vec{a}^{(k)} - \vec{a}\|_\infty \leq \epsilon^\prime + {\frac{U}{\tau+\eta}}.
\end{equation*}
{ \it Furthermore, for $\epsilon_c { \leq } 2 \log \left ( {\vec{a}_{\rm max}}/{\vec{a}_{\rm min}} \right)$, taking $\tau = ({2\|\mat{C}\|_\infty})/{\epsilon_c} + \eta(\frac{2}{\epsilon_c}\log \left ( {\vec{a}_{\rm max}}/{\vec{a}_{\rm min}} \right) - 1), \forall {\eta >0}$, $\epsilon^\prime = \frac{\epsilon_c}{2}$, $\|\vec{a}^{(k)} - \vec{a}^* \|_\infty \leq \epsilon_c$ holds. Otherwise, for $\epsilon_c { \geq } 2 \log \left ( {\vec{a}_{\rm max}}/{\vec{a}_{\rm min}} \right)$, taking $\tau = (2\|\mat{C}\|_\infty)/\epsilon_c + \eta(\frac{2}{\epsilon_c}\log \left ( {\vec{a}_{\rm max}}/{\vec{a}_{\rm min}} \right) - 1), \forall \eta, 0 \leq \eta \leq (2\|\mat{C}\|_\infty) / (\epsilon_c(1 - \frac{2}{\epsilon_c}\log \left ( {\vec{a}_{\rm max}}/{\vec{a}_{\rm min}} \right) ))$, and $\epsilon^\prime = \frac{2}{\epsilon_c}$, then $\|\vec{a}^{(k)} - \vec{a}^* \|_\infty \leq \epsilon_c$ holds.}

We now provide the proof of  {\bf Theorem \ref{thm:AlgorithmTimeMarginalComplexity}}.
\begin{proof}
{Considering the stopping criterion of {\bf Lemma \ref{lem:NewStoppingCreterion}} and the upper bound of {\bf Lemma \ref{lem:new_supreme_norm}} under the condition where $\alpha = \beta = 1$}, the relaxed marginal constraint {gap} is provided as
\begin{equation*}
	\|\vec{a}^{(k)} - \vec{a}\|_\infty \leq \gamma \left ( \frac{4\tau}{\eta}R \left ( \frac{\tau}{\tau+\eta} \right)^{{\frac{k-1}{2}-1}} + \frac{\|\vec{u}^*\|_\infty}{\tau} \right) \leq \epsilon^\prime + \frac{U}{\tau+\eta},
\end{equation*}
where $U$ represents $\|{\bf C}\|_\infty + \eta (\log \vec{a}_{\rm max} - \log \vec{a}_{\rm min})$.

{We then consider the selection of $\tau$ and $\eta$. When taking $\epsilon_c$ satisfying $\epsilon_c \leq 2 (\log \vec{a}_{\rm max} - \log \vec{a}_{\rm min})$ and defining $\tau = \frac{2\|{\bf C} \|_\infty}{\epsilon_c} + \eta (\frac{2}{\epsilon_c}(\log \vec{a}_{\rm max} - \log \vec{a}_{\rm min}) -1)$ for $\eta > 0$,  we can keep $\tau$ positive. Thus, defining $\epsilon^\prime = \frac{\epsilon}{2}$, $\|\vec{a}^{(k)} - \vec{a}\|_\infty$ holds. }

Otherwise, taking $\epsilon_c$ satisfying $\epsilon_c \geq 2 (\log \vec{a}_{\rm max} - \log \vec{a}_{\rm min})$ and defining $\tau = \frac{2\|{\bf C} \|_\infty}{\epsilon_c} + \eta (\frac{2}{\epsilon_c}(\log \vec{a}_{\rm max} - \log \vec{a}_{\rm min}) -1)$ for $ 0 < \eta <  (2\|{\bf C}\|_\infty)/(\epsilon_c - 2(\log \vec{a}_{\rm max} - \log \vec{a}_{\rm min}))$, we also can keep $\tau$. Likewise, defing $\epsilon^\prime = \frac{\epsilon}{2}$, $\|\vec{a}^{(k)} - \vec{a}\|_\infty$ holds. This completes the proof.
\end{proof}

\section{Theoretical results about the OT distance {gap}}
{This section provides the proof of {\bf Theorem \ref{thm:approximation_originalOT}}.}
{For that, we first redescribe {\bf Theorem \ref{thm:approximation_originalOT}} in the main material}. 

\noindent{{\bf Theorem 4.11} (convergence to $\epsilon$-approximation of OT distance gap).
{\it Letting $\mat{T}^{\rm OT}$, $c_3$ and $U$ be the optimal solution of {the standard OT problem (\ref{eq:FormulationOptimalTransport})}, $2\log n  + 1 - \defmax \lbrace \mathrm{H}(\vec{a}), \mathrm{H}(\vec{b}) \rbrace)$ and {$ \|\mat{C}\|_\infty+\eta \log \left( {\vec{a}_{\rm max}}/{\vec{a}_{\rm min}}\right)$, respectively}. Assume $\vec{a},\vec{b} \in \Delta^{n}$. Also, one considers the case in which $k$ is even after the odd update. Let $\mat{T}^{(k)}$ and $\mat{Y}$ be the matrix generated by the SR--Sinkhorn algorithm and {its projected matrix by \cite[Algorithm 2]{altschuler2017nearlinear}, respectively}. Then, the OT distance gap is provided as}
\begin{equation*}
	\langle \mat{C},\mat{Y} \rangle - \langle \mat{C},\mat{T}^{\mathrm{OT}} \rangle \leq (2n\|\mat{C}\|_{\infty}+\|\mat{C}\|_1)\frac{4\tau}{\eta}R\left (\frac{\tau}{\tau + \eta}\right )^{ \frac{{k-1}}{2}  -1} +\eta c_3 + \frac{2n\|C\|_{\infty}}{\tau}U.
\end{equation*}
{\it In addition, if $\|\log \left ( \frac{{\bf T}^{(k)}}{{\bf T}^{*}} \right) \|_\infty \leq \epsilon^{\prime}$, {$\mat{Y}$} satisfies}
\begin{equation*}
	\langle \mat{C},{\mat{Y}} \rangle - \langle \mat{C}, \mat{T}^{\rm OT} \rangle \leq (2n\|\mat{C}\|_{\infty}+\|\mat{C}\|_1)  \epsilon^\prime  + \eta c_3 + \frac{2n\|\mat{C}\|_\infty}{\tau}U.
\end{equation*}
{\it Defining $\epsilon^\prime = \frac{\epsilon_d}{{3(2n\|{\bf C}\|_\infty + \|{\bf C}\|_1)}}$, $\eta = \frac{\epsilon_d}{3c_3}$ and $\tau = \frac{6n\|{\bf C}\|_\infty}{{\epsilon_d}}U$, $\langle \mat{C},\mat{Y} \rangle - \langle \mat{C}, \mat{T}^{\rm OT} \rangle \leq \epsilon_d$ holds.} }

Before giving the proof, we describe the projection algorithm onto the the domain of the standard OT {\cite[Algorithm 2]{altschuler2017nearlinear}} and the relevant inequality.
\begin{algorithm}[htbp]
\caption{Projection onto the domain of the standard OT \cite[Algorithm 2 ]{altschuler2017nearlinear}}      
\label{Appenalg:ProjectionOntotheOT}    
\begin{algorithmic}[1]
\REQUIRE{$\mat{X} \in \mathbb{R}^{n \times n}$, $\vec{a}, \vec{b} \in \mathbb{R}^n$}
\ENSURE{$\mat{Y}$}

\STATE {$\mat{P} = \mathrm{diag}(\vec{x})$ with $\vec{x}_i = \min \left (\frac{\vec{a}_i}{\displaystyle{({{\bf X}}\bm{1}_n)_i}}, 1\right )$} 
\STATE {$\mat{X}^\prime = \mat{P}\mat{X}$}
\STATE {$\mat{Q} = \mathrm{diag}(\vec{y})$ with $\vec{y}_i = \min \left (\frac{\vec{b}_i}{\displaystyle{({\mat{X}^{\prime}}^T\vec{1}_n)_j}}, 1\right )$} 
\STATE {$\mat{X}^{\prime\prime} = \mat{X}^{\prime}\mat{Q}$}
\STATE{$err_r = \vec{a} - \mat{X}^{\prime\prime},err_c = \vec{b} - {\mat{X}^{\prime\prime}}^T\vec{1}_n$}
\STATE{$ \mat{Y} = \mat{X}^{\prime\prime} + err_r err_c^T/\|err_r\|_1$}

\end{algorithmic}
\end{algorithm}

\begin{Lem}(Lemma 7 \cite{altschuler2017nearlinear}) For the inputs $\mat{X} \in \mathbb{R}^{n \times n},\vec{a}, \vec{b} \in \mathbb{R}^n$, {\bf Algorithm \ref{Appenalg:ProjectionOntotheOT} } takes $\mathcal{O}(n^2)$ time to output a matrix $\mat{Y} \in \mathcal{U}(\vec{a},\vec{b})$ satisfying:
\label{Appenlem:OutputInequality}
\begin{equation*}
\|\mat{Y} - \mat{X}\|_1 \leq 2 \left ( \|\mat{X}\vec{1}_n - \vec{a}\|_1 + \|\mat{X}^T\vec{1}_n - \vec{b}\|_1 \right)
\end{equation*}
\end{Lem}

We now provide the proof of {\bf Theorem \ref{thm:approximation_originalOT}}. Hereinafter, we denote $\mat{T}^{\rm OT}$ and $\mat{Y}$ as the optimal solution of (\ref{eq:FormulationOptimalTransport}) and a matrix computed by {\bf Algorithm \ref{Appenalg:ProjectionOntotheOT}} for $\mat{T}^{(k)}$, respectively.
\begin{proof}
We have
\begin{eqnarray*}
\langle \mat{C} ,  \mat{Y} \rangle - \langle \mat{C}, \mat{T}^{\rm OT} \rangle &=&\langle \mat{C}, \mat{Y} - \mat{T}^{(k)} \rangle + \langle \mat{C}, \mat{T}^{(k)}-\mat{T}^* \rangle + \langle \mat{C}, \mat{T}^* - \mat{T}^{\rm OT} \rangle \\
&\leq& \|\mat{C}\|_{\infty}\|\mat{Y} - \mat{T}^{(k)}\|_1 + \|\mat{C}\|_1\|\mat{T}^{(k)}-\mat{T}^*\|_\infty + \langle \mat{C}, \mat{T}^* - \mat{T}^{\rm OT} \rangle,
\end{eqnarray*}
{where the {first inequality} uses {the} Holder's inequality. These three terms can be bounded separately as given below.}

\noindent {\bf Upper-bound of $\|\mat{T}^{(k)} - \mat{T}^{*}\|_\infty$} : From {\bf Lemma \ref{lem:NewStoppingCreterion}} and {\bf Lemma \ref{lem:KeyInequality}}, this term is bounded as
\begin{equation}
\label{eq:ub_t_t}
	\|\mat{T}^{(k)} - \mat{T}^{*}\|_\infty \leq \|\log \mat{T}^{(k)} - \log \mat{T}^{*}\|_\infty \leq \frac{4\tau}{\eta}R\left (\frac{\tau}{\tau+\eta}\right )^{{ \frac{k-1}{2} } -1} \leq \epsilon^\prime.
\end{equation}

\noindent {\bf Upper-bound of $\|\mat{Y} - \mat{T}^{(k)}\|_1$}: Because $\mat{Y}$ is generated by {\bf Algorithm \ref{Appenalg:ProjectionOntotheOT} }, we obtain from {\bf Lemma \ref{Appenlem:OutputInequality}}
\begin{eqnarray*}
	\|\mat{Y} - \mat{T}^{(k)}\|_1 &\leq& 2 \left (\|\mat{T}^{(k)}\vec{1}_n - \vec{a}\|_1 + \|{\mat{T}^{(k)}}^T\vec{1}_n- \vec{b}\|_1 \right ) \\
	&{=}& 2 n \|\mat{T}^{(k)}\vec{1}_n - \vec{a}\|_\infty \\
	&\leq& 2n \left ( \frac{4\tau}{\eta}R\left(\frac{\tau}{\tau + \eta}\right )^{ \frac{{k-1}}{2}  -1} +  \frac{U}{\tau+\eta} \right )\\
	&\leq& 2n \left(\epsilon^\prime + \frac{U}{\tau} \right),
\end{eqnarray*}
where the {second} inequality uses {\bf Theorem \ref{thm:AlgorithmTimeMarginalComplexity}}, and the {last inequality uses (\ref{eq:ub_t_t})} as the same as the case of the upper bound of $\|\mat{T}^{(k)} - \mat{T}^{*}\|_\infty$.

\noindent {\bf Upper-bound of $\langle \mat{C}, \mat{T}^{*} -  \mat{T}^{\rm OT}  \rangle $} : We are inspired by the proof of \cite[{{\bf Theorem 4.3}}]{extrapolation2022Quang}. We have the following inequality for $g(\mat{T}^{\rm OT})$ where $\mat{T}^{\rm OT} \in \mathcal{U}(\vec{a},\vec{b})$:
\begin{eqnarray*}
	g(\mat{T}^{\rm OT}) &=& \langle \mat{C} , \mat{T}^{\rm OT} \rangle + \tau \mathrm{KL}(\mat{T}^{\rm OT}\vec{1}_n,\vec{a})- \eta \mathrm{H}(\mat{T}^{\rm OT}) \\
	&\leq &  \langle \mat{C} , \mat{T}^{\rm OT} \rangle -\eta (\max \lbrace \mathrm{H}(\vec{a}),\mathrm{H}(\vec{b}) \rbrace) ,
\end{eqnarray*}
where we use $\mathrm{KL}((\mat{T}^{\rm OT}\vec{1}_n,\vec{a}) = 0$, and the upper bound of the entropy term in \cite{blondel2018smooth}. 

{As for $g(\mat{T}^{*})$,} we have
\begin{eqnarray*}
	g(\mat{T}^{*}) &=& \langle \mat{C}, \mat{T}^{*} \rangle + \tau \mathrm{KL}(\mat{T}^{*}\vec{1}_n,\vec{a}) - \eta \mathrm{H}(\mat{T}^{*}) \\
	&\geq& \langle \mat{C}, \mat{T}^{*} \rangle  - \eta \mathrm{H}(\mat{T}^{*})\\
	&\geq& \langle \mat{C}, \mat{T}^{*} \rangle - \eta (2\log n + 1),
\end{eqnarray*}
{where we used the fact that the KL divergence is non-negative, and where the last inequality uses} $\beta=1$ for $2\beta \log n + \beta - \beta \log \beta$ \cite{UOTSinkhorn2020}.
{Here, addressing $\mat{T}^{*} = \defargmin_{{\bf T} \geq {\bf 0},{\bf  T}^T\bm{1}_n = \bm{b}} g(\mat{T})$, we have $g(\mat{T}^{\rm OT}) \geq g (\mat{T}^{*})$. Therefore, we obtain}
\begin{equation}
	\langle \mat{C}, \mat{T}^{*} \rangle - \langle \mat{C} , \mat{T}^{\rm OT} \rangle \leq \eta (2\log n + 1 - \max \lbrace \mathrm{H}(\vec{a}),\mathrm{H}(\vec{b}) \rbrace ):=\eta c_3.
\end{equation}

Finally, putting all these three ingredients together, we can obtain
\begin{equation}
	\langle \mat{C}, \mat{Y} \rangle - \langle \mat{C}, \mat{T}^{\rm OT} \rangle \leq  (2n\|\mat{C}\|_\infty + \|\mat{C}\|_1)\epsilon^\prime + \frac{2n\|\mat{C}\|_\infty}{\tau}U + \eta c_3
\end{equation}

Taking $\epsilon^\prime = \frac{\epsilon_d}{3(2n + \|{\bf C}\|_\infty + \|{\bf C}\|_1)}$,  $\eta = \frac{\epsilon_d}{3c_3}$ and $\tau = \frac{6n\|{\bf C}\|_\infty}{\epsilon_d}U$, $\langle \mat{C},\mat{Y}\rangle - \langle \mat{C}, \mat{T}^{\rm OT} \rangle \leq \epsilon^\prime_d$ holds. 
\end{proof}

\section{Marginal constraint gap bound of the SROT}
In this section, we provide a new theoretical result of the upper bound of the marginal constraint gap of the SROT {without the entropy regularization, which is defined in  (\ref{eq:KLSemiRelaxed})}. Herein, we assume that $\vec{a},\vec{b} \in \Delta^{{n}}$. First, we consider the following lemma.
\begin{Lem}{}
\label{lem:marginal_bound_semi_entropy}
Let $\mat{T}^*$ be the optimal solution of (\ref{eq:EntropySemiRelaxed}). Then, $\mat{T}^*$ satisfies
\begin{equation*}
	\|\mat{T}^*\vec{1}_n - \vec{a} \|_1 \leq \frac{nU}{\tau},
\end{equation*}
where $U = \frac{\tau}{\tau + \eta} \left ({\|\mat{C}\|_\infty} + \eta  ( \log \vec{a}_{\rm max} - \log \vec{a}_{\rm min})\right )$.
\end{Lem}
\begin{proof}
We refer to the proof of {\cite[{{\bf Lemma E.3}}]{extrapolation2022Quang}}.
Considering  $U = \frac{\tau}{\tau + \eta} \left ({\|\mat{C}\|_\infty} + \eta  ( \log \vec{a}_{\rm max} - \log \vec{a}_{\rm min})\right )$, we can similarly prove (\ref{lem:marginal_bound_semi_entropy}).
\end{proof}

\begin{Thm}[The relaxed marginal constraint bound]
Let $\hat{\mat{T}}$ be the optimal solution of (\ref{eq:KLSemiRelaxed}). Assume that $\vec{a} \in \Delta^n$. Then, this solution satisfies
\begin{equation}
	\label{eq:marginal_bound_semi_relaxed}
	\|\hat{\mat{T}}\vec{1}_n - \vec{a}\|_1 \leq \frac{n\|\mat{C}\|_\infty}{\tau}.
\end{equation}
\end{Thm}
\begin{proof}
{By following} the proof of {\cite[{\bf Theorem 4.2}]{extrapolation2022Quang}}, we use the Bolzano-Weierstrass because $\|\mat{T}^*\|_1 = 1$ is the bounded close set. Therefore, there exists a convergent subsequence $ \lbrace \mat{T}^{\eta_{t}} \rbrace^{\infty}_{{t=}1}$ {such that} $\lim_{t \to \infty} {\eta_t = 0}$, where $\mat{T}^{\eta_{t}}$ is the optimal solution of (\ref{eq:EntropySemiRelaxed}) for regularization parameter $\eta_t$. {In addition, we can} define the limit of this subsequence as $\hat{\mat{T}} = \lim_{\eta_t \to 0} \mat{T}^{\eta_t}$ {because $\hat{\mat{T}}$ satisfies the constraint $\|\hat{\mat{T}}\|_1 = 1$} . We first prove 
\begin{equation*}
	\lim_{\eta_t \to 0} \langle \mat{T}^{{\eta_t}},\mat{C} \rangle + \tau\mathrm{KL}(\mat{T}^{{\eta_t}}\vec{1}_n,\vec{a})-\eta_t \mathrm{H}(\mat{T}^{{\eta_t}}) =\langle \hat{\mat{T}},\mat{C} \rangle + \tau\mathrm{KL}(\hat{\mat{T}}\vec{1}_n,\vec{a})=f(\hat{\mat{T}}).
\end{equation*}
We obtain that
\begin{eqnarray*}
	\lim_{\eta_t \to 0} \langle \mat{T}^{\eta_t},\mat{C} \rangle + \tau\mathrm{KL}(\mat{T}^{\eta_t}\vec{1}_n,\vec{a})-\eta_t \mathrm{H}(\mat{T}^{\eta_t}) 
	&=& \lim_{\eta_t \to 0} f(\mat{T}^{\eta_t}) - \eta_t \mathrm{H}(\mat{T}^{\eta_t}) \\
	&=& \lim_{\eta_t \to 0} f(\mat{T}^{\eta_t}) \\
	&=& f (\lim_{\eta_t \to 0} \mat{T}^{\eta_t})\\
	&=& f(\hat{\mat{T}}),
\end{eqnarray*}
where the second {equality} comes from $0 \leq \mathrm{H}(\mat{T}^{\eta_t}) \leq 2\log n + 1$ and the last {equality} is from the continuity of the function $f$. From {\bf Lemma \ref{lem:marginal_bound_semi_entropy}}, we have
\begin{equation*}
	\|\mat{T}^{\eta_t}\vec{1}_n - \vec{a} \|_1 \leq  \frac{n}{\tau + \eta_t} \left ({\|\mat{C}\|_\infty} + \eta_t  ( \log \vec{a}_{\rm max} - \log \vec{a}_{\rm min})\right ).
\end{equation*}
Considering $\eta_t \to {0}$, we can obtain the bound (\ref{eq:marginal_bound_semi_relaxed}).
\end{proof}

\section{Additional numerical evaluations}
{This section provides additional numerical evaluations. It has two purposes: The first one is to evaluate the obtained theoretical iteration bound by comparing the empirical numerical results. Second, we will show that the SR-Sinkhorn possesses intermediate properties of convergence between the original Sinkhorn algorithm and the fully relaxed algorithm. In addition, the proposed algorithm is comparable with the state-of-the-art algorithms dedicated to the SROT problem. }

\subsection{Comparisons of theoretical computational complexity with {\cite{NeurIPS2021robust}}}
{We analyze the theoretical stopping iteration bound in {\bf {Corollary} \ref{cor:ComplexityAndOrder}} of the SR--Sinkhorn algorithm in comparison with that of \cite{NeurIPS2021robust}. We also compare them with the theoretical result of the} UOT--Sinkhorn algorithm \cite{UOTSinkhorn2020}. For this purpose, by following \cite{UOTSinkhorn2020}, two stopping iteration numbers $k$ are calculated: Given approximation constants $\epsilon$ selected uniformly from {$1.0$ to $0.05$}, the first $k_f$ is calculated from the stopping criterion obtained in {\bf Corollary \ref{cor:ComplexityAndOrder}} according to $\epsilon$.  $k_c$ is, on the other hand, the second, the measured value in the numerical experiments, which is the smallest value satisfying $|f(\mat{T}^{(k)}) - f(\hat{\mat{T}})| \leq \epsilon$. Also, $\eta$ is automatically set as $\eta = \frac{\epsilon}{2c_2}$ in {\bf Theorem \ref{thm:AlgorithmTimeComplexity}}.

As for a dataset, after setting {$n=100$} and {$\tau={1,10,100}$}, the elements of the cost matrix $\mat{C}$ are generated uniformly from the closed interval {$[1,100]$}. The weight vectors $\vec{a},\vec{b}$ are also configured uniformly from the closed interval {$[1,10]$} and normalized to $1$ respectively. 

\begin{figure}[p]
	\begin{minipage}[t]{\textwidth}
	\begin{center}
	\includegraphics[width=0.8\textwidth]{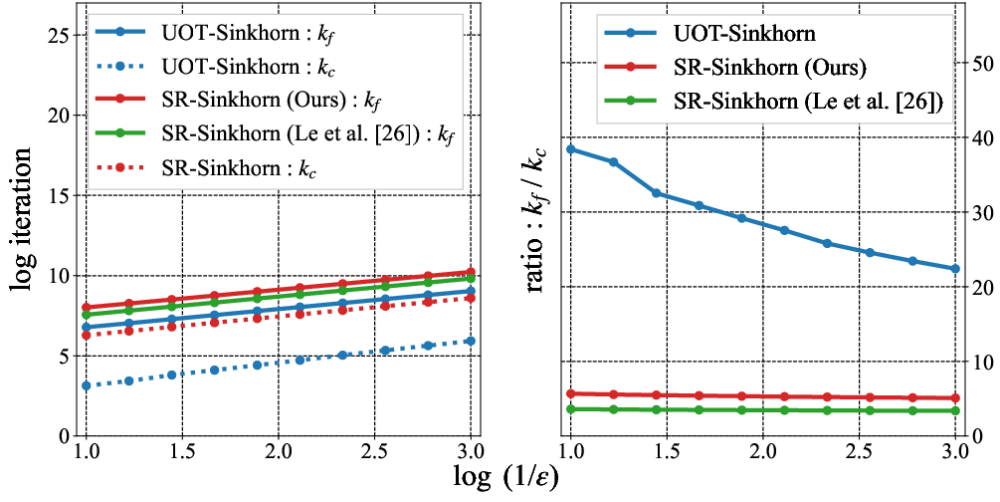}\\
	\vspace*{0.1cm}
	{\large (a) $\tau = 1.0 \times 10^0$.}
	\vspace*{0.1cm}
	\end{center}
	\end{minipage}

	\begin{minipage}[t]{\textwidth}
	\begin{center}
	\includegraphics[width=0.8\textwidth]{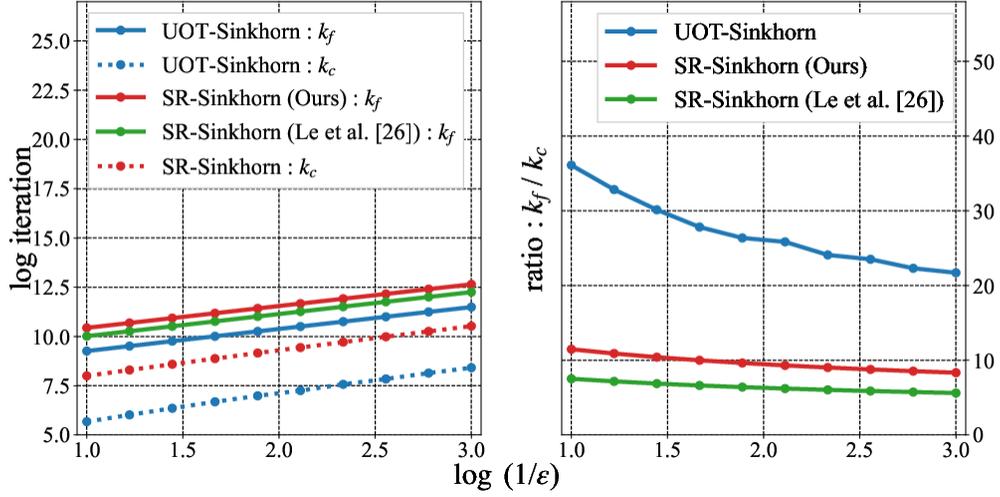}\\
	\vspace*{0.1cm}
	{\large (b) $\tau = 1.0 \times 10^1$.}
	\vspace*{0.1cm}
	\end{center}
	\end{minipage}
	\begin{minipage}[t]{\textwidth}
	\begin{center}
	\includegraphics[width=0.8\textwidth]{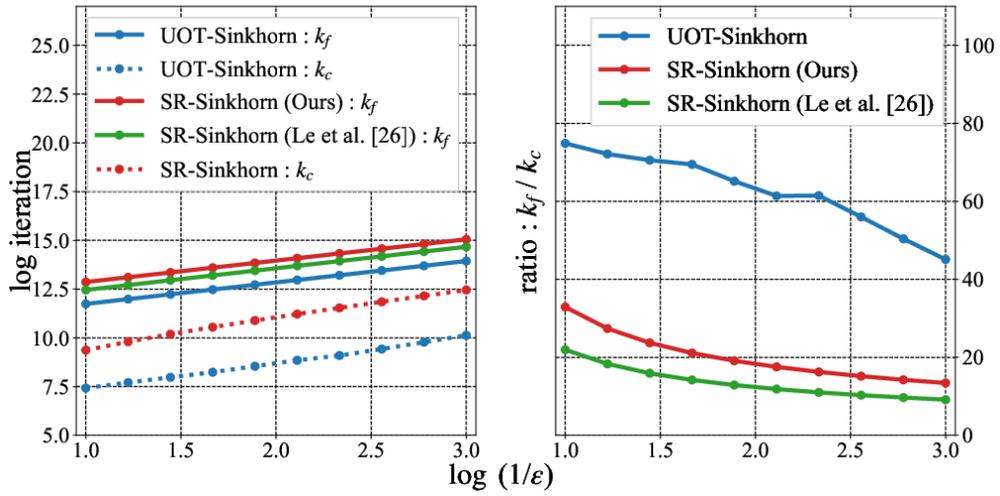}\\
	\vspace*{0.1cm}
	{\large (c) $\tau = 1.0 \times 10^2$.}
	\vspace*{0.1cm}
	\end{center}
	\end{minipage}
	
	\caption{Comparison theoretical and empirical iterations. Left: the convergence logarithmic iterations $k_f, k_c$. Right: the ratio $k_f/k_c$}	
	\label{fig:ComparisonIteration}		
\end{figure}

Figure \ref{fig:ComparisonIteration} shows the log iteration of $k_f$ and $f_c$, and the ratio $k_f / f_c$ for the three algorithms. {Note that this includes one single result of $k_c$ for the SR--Sinkhorn algorithm because $k_c$ is the numerical results and, ours and \cite{NeurIPS2021robust} are the same.} From this figure, as one might expect, $k_f$ and $k_c$ {of ours and \cite{NeurIPS2021robust}} are larger than those of the UOT--Sinkhorn because the UOT--Sinkhorn is more relaxed than our formulation is. It is, however, noteworthy that the ratio $k_f/k_c$ for all $\tau$ {of ours and \cite{NeurIPS2021robust} indicate} much smaller  values than that of the UOT--Sinkhorn. This finding implies that {both bounds of the SR--Sinkhorn algorithm are} is closer to the practical iteration, and they are tighter than that of the UOT-Sinkhorn. {In comparison with theoretical result of \cite{NeurIPS2021robust},} {our log iteration and the ratio indicate slightly larger values than those of \cite{NeurIPS2021robust}, respectively. However, as seen in the figure, the differences are reasonably small, and those behaviors resemble. This observation coincides with the fact that our obtained complexity is the same order as that of \cite{NeurIPS2021robust} except constant numbers.} 

\begin{figure}[p]
	\begin{minipage}[t]{\textwidth}
	\begin{center}
	\includegraphics[width=0.7\textwidth]{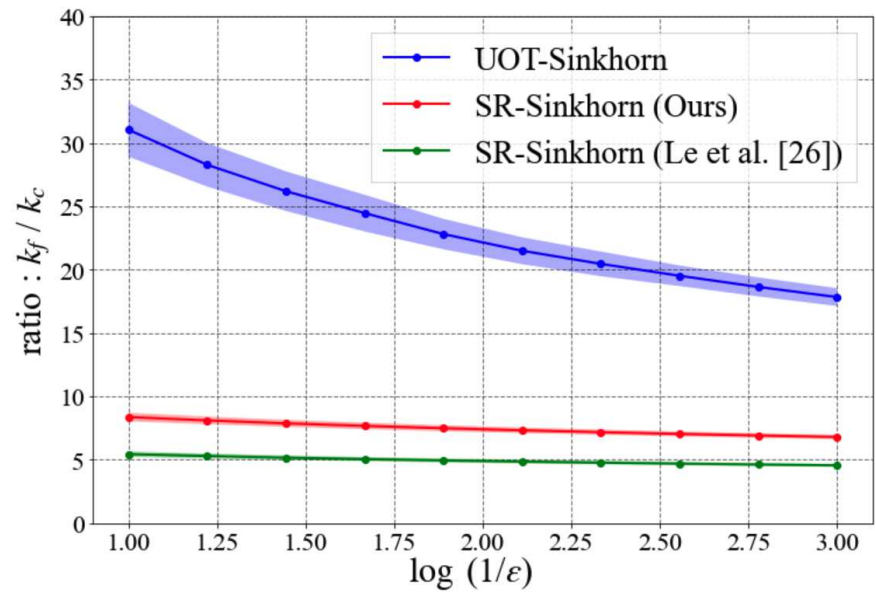}\\
	\end{center}
	\end{minipage}
	\caption{Comparison theoretical and empirical iterations with the mean and the standard deviation.}	
	\label{fig:ComparisonIterationWithMean}		
\end{figure}

Next, we run this experiment $10$ times and compute their means and standard deviation value in Figure \ref{fig:ComparisonIterationWithMean}. This experimental settings are the same conditions as {the previous one except for $\tau = 5$}.  From this figure, we can understand that the standard deviations of {ours and \cite{NeurIPS2021robust} are also smaller than that of UOT. Specifically, the standard deviation of ours and \cite{NeurIPS2021robust} approaches $0$ as $\epsilon$ gets smaller. Similarly to the previous experiments, we can see some differences between ours and that of \cite{NeurIPS2021robust}, but they are reasonably small, and those behaviors resemble.}

{Overall, the theoretical convergence results in terms of the functional gap in our proof are slightly worser than that of \cite{NeurIPS2021robust}, but those degradations are not so larger. Rather, it should be emphasized that this similar result of the the functional gap is derived from our new new proof strategy, and this proof derives new additional but important theoretical results: the marginal constraint gap and the OT distance gap. They have not been addressed in the literature of the SROT problem. Those analyses are in the main material, and the numerical results support our theoretical results.}

\clearpage

\subsection{ Comparison with Sinkhorn algorithms.} 
\begin{figure}[h]
\begin{center}
	\begin{minipage}[t]{0.49\textwidth}
	\begin{center}
	\includegraphics[width=\textwidth]{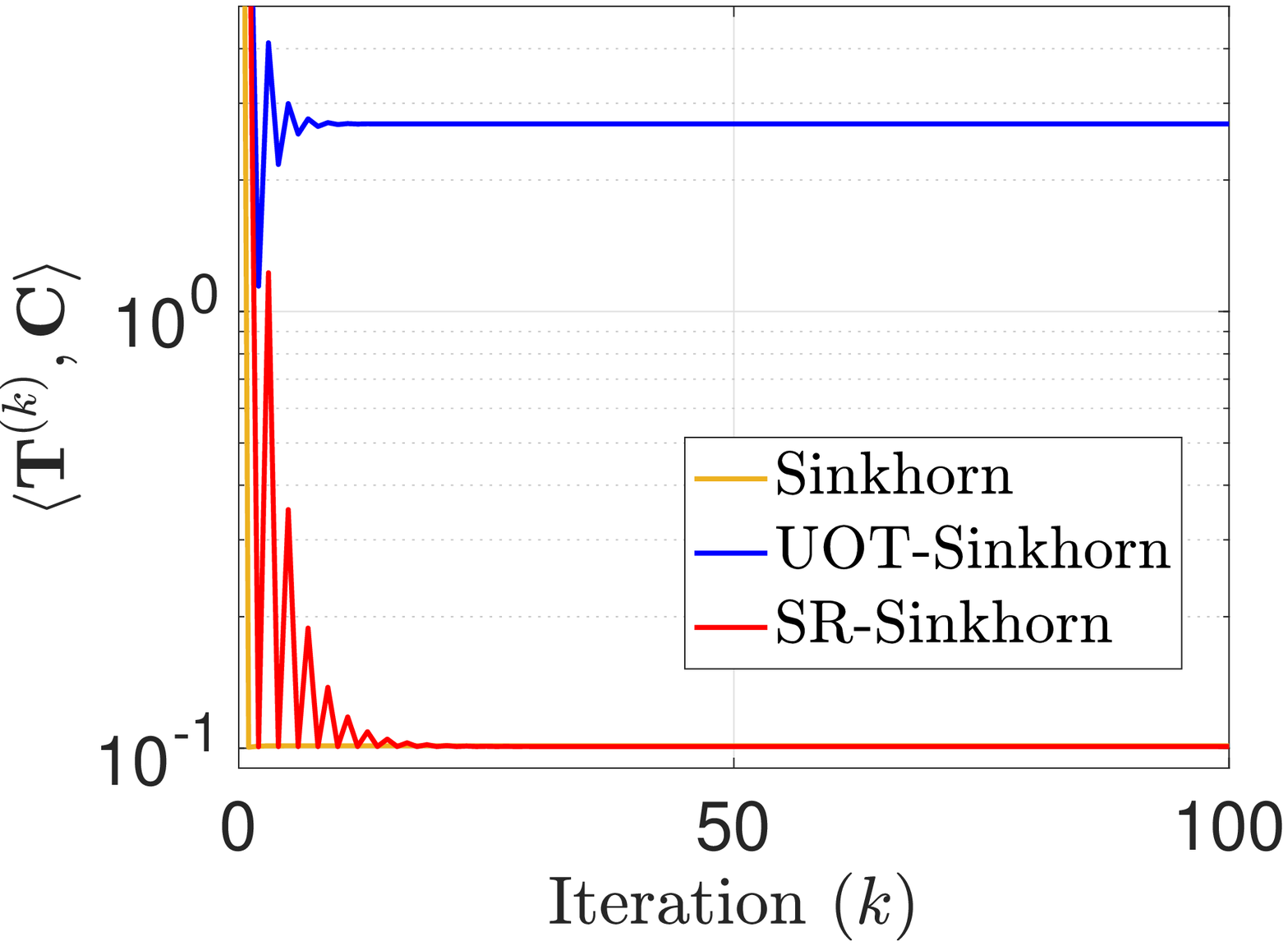}	
	
	{\small (a) $\langle \mat{T}^{(k)}, \mat{C}\rangle$.}
	\end{center}
	\end{minipage}
	\hspace*{-0.4cm}	
	\begin{minipage}[t]{0.49\textwidth}
	\begin{center}	
	\includegraphics[width=\textwidth]{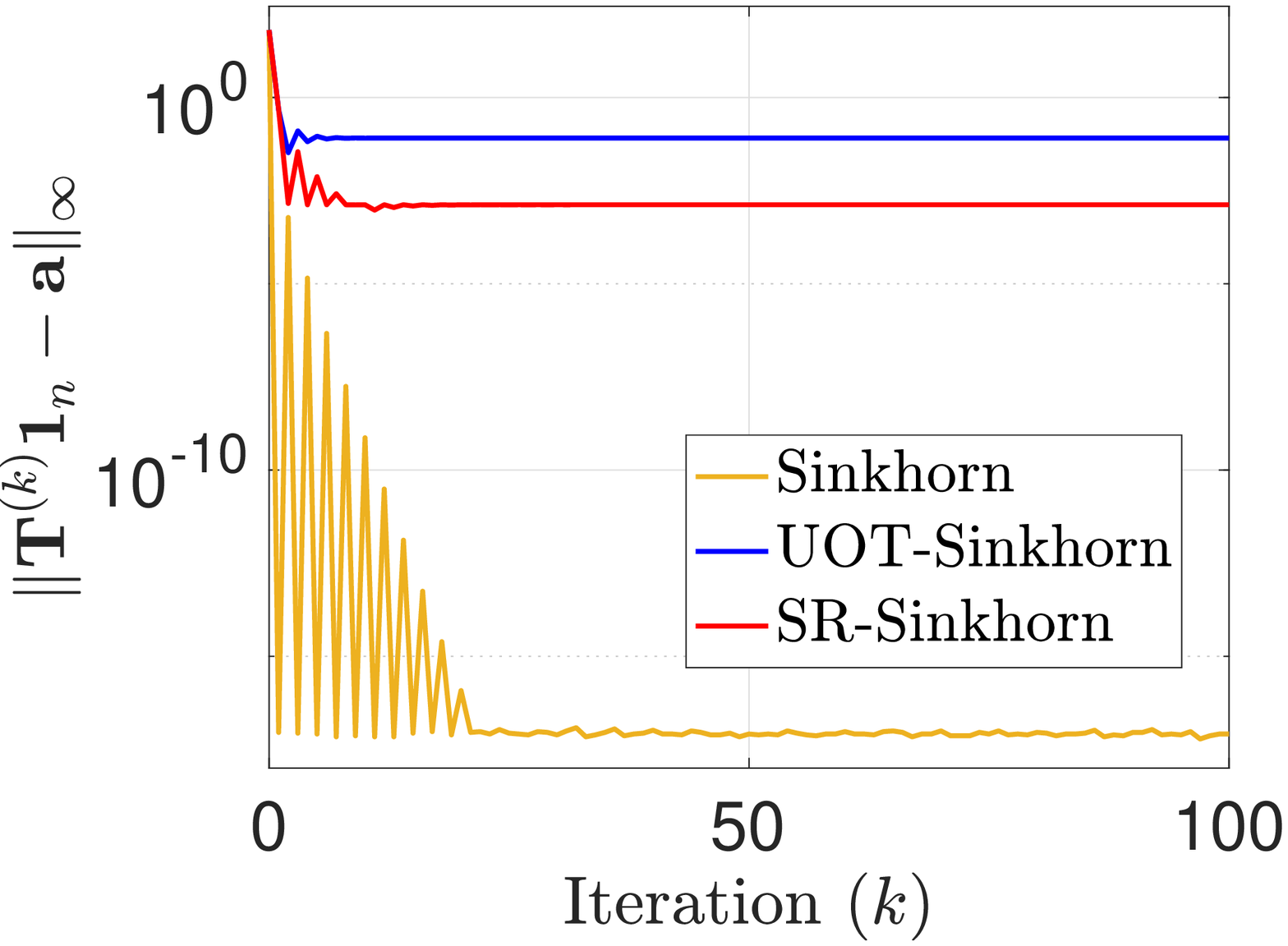}	
	
	{\small (b) $\|\mat{T}^{(k)}\!\vec{1}_n\!\! - \!\vec{a}\|_{\!\infty}$.}
	\end{center}	
	\end{minipage}	
	\hspace*{-0.4cm}	
	\begin{minipage}[t]{0.49\textwidth}
	\begin{center}	
	\includegraphics[width=\textwidth]{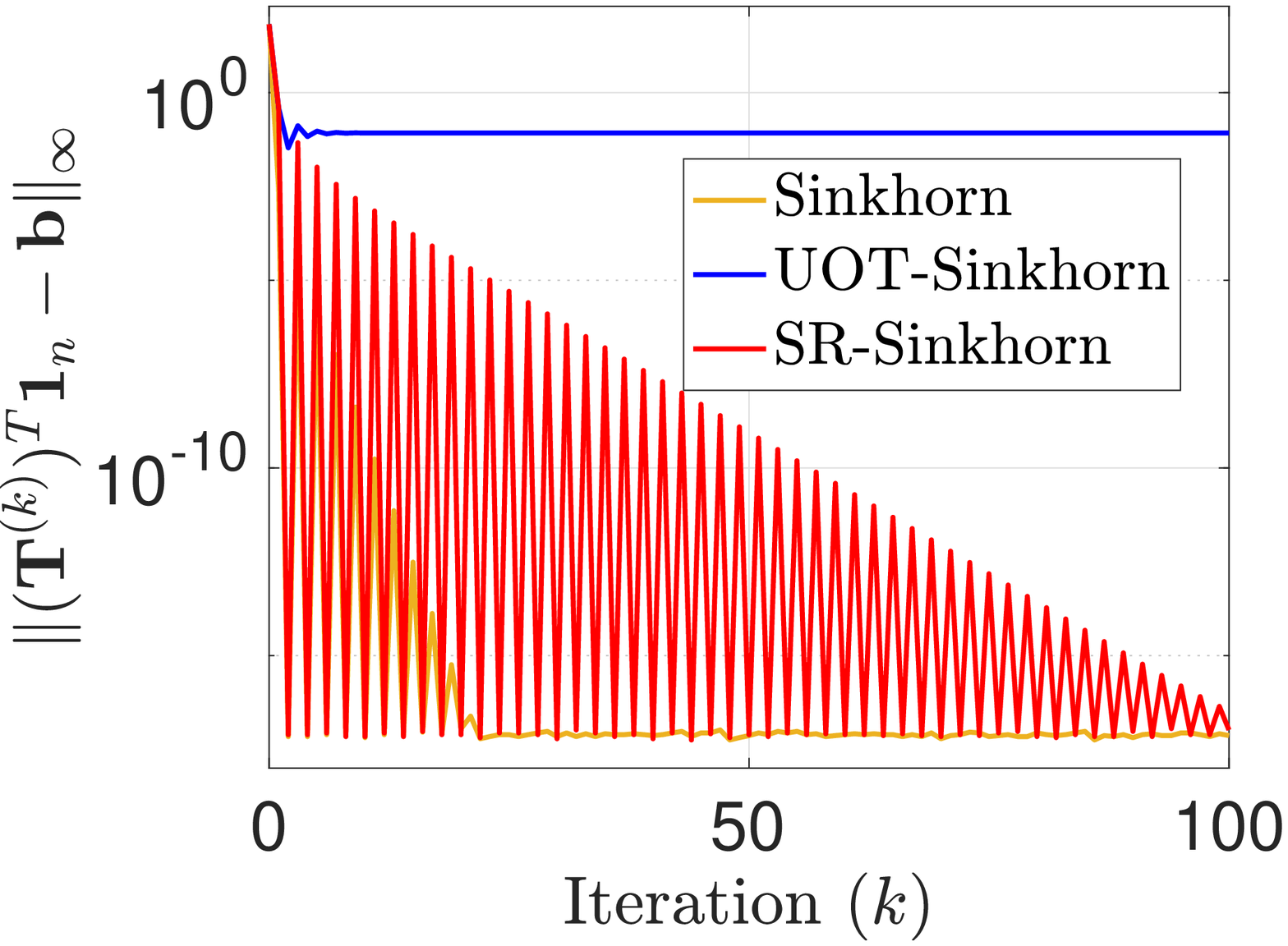}	
	
	{\small (c) $\|(\mat{T}^{(k)})^T\!\vec{1}_n \!-\! \vec{b}\|_\infty$.}
	\end{center}	
	\end{minipage}	
	
	\caption{Comparison with Sinkhorn algorithms.}	
	\label{fig:Sinkhorn_Comparison}	
\end{center}		
\end{figure}

To reveal fundamental characteristics of the SR--Sinkhorn, we experimentally measure, at every iteration $k$, the distance $\langle \mat{T}^{(k)}, \mat{C}\rangle$ and the marginal gaps of $\|\mat{T}^{(k)}\vec{1}_n - \vec{a}\|_\infty$ and $\|(\mat{T}^{(k)})^T\vec{1}_n - \vec{b}\|_\infty$. Figure \ref{fig:Sinkhorn_Comparison} shows these three values in (a), (b), and (c), respectively. For this experiment, we generate a synthetic dataset of size $n=500$, and execute the Sinkhorn, UOT--Sinkhorn \cite{UOTSinkhorn2020}, and SR--Sinkhorn algorithms under $(\eta, \tau, \tau_1 ,\tau_2)=(0.1,0.1,0.1,0.1)$. The maximum iteration number is $100$. These graphs indicate that the SR--Sinkhorn algorithm possesses intermediate properties of convergence between the Sinkhorn and UOT--Sinkhorn algorithms. 

\subsection{Comparison with SROT algorithms.} 
\begin{figure}[h]
\begin{center}
	\begin{minipage}[t]{0.49\textwidth}
	\begin{center}
	\includegraphics[width=\textwidth]{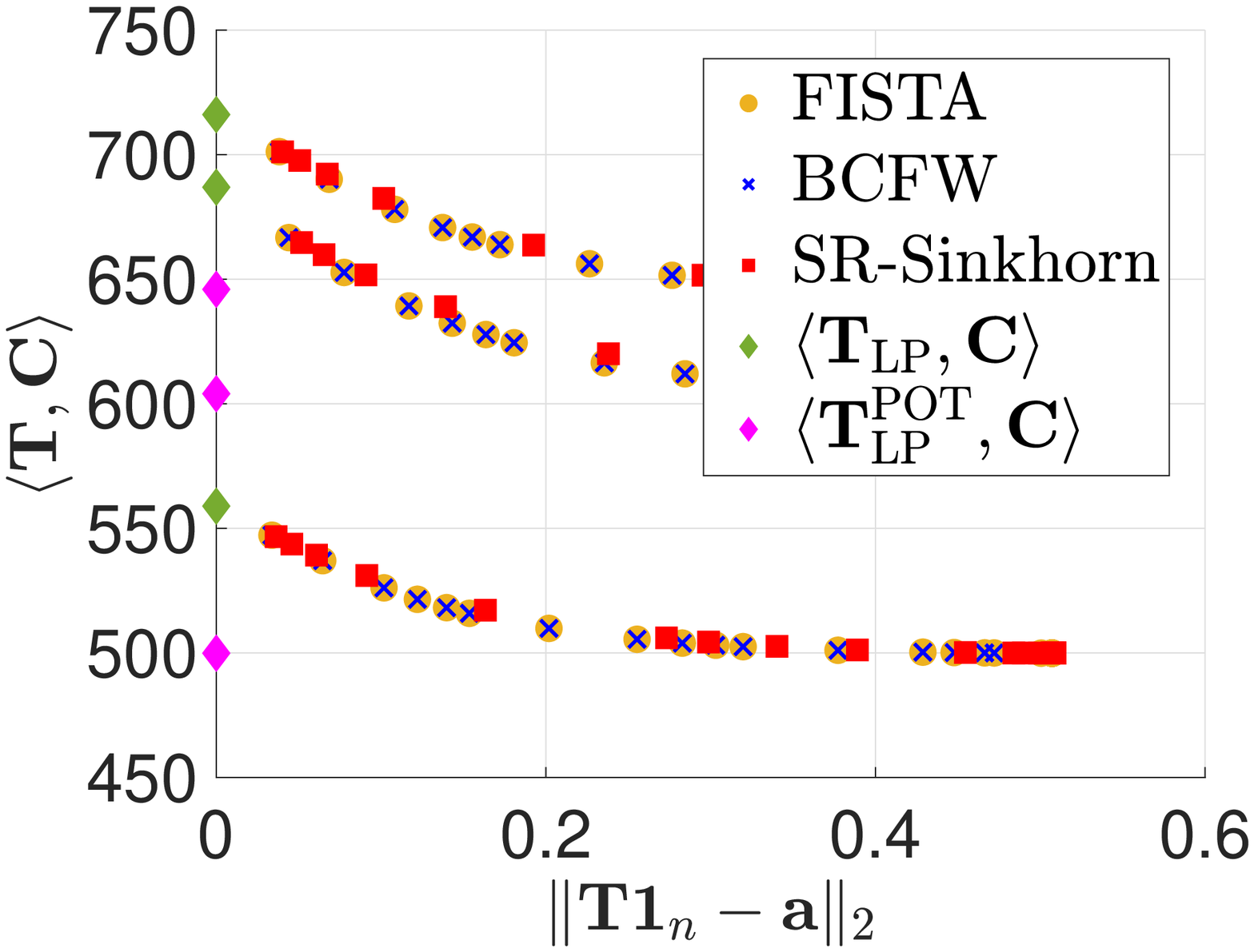}	
	
	{\small (a) distance: $\langle \mat{T}, \mat{C}\rangle$.}
	\end{center}	
	\end{minipage}
	\begin{minipage}[t]{0.49\textwidth}
	\begin{center}
	\includegraphics[width=\textwidth]{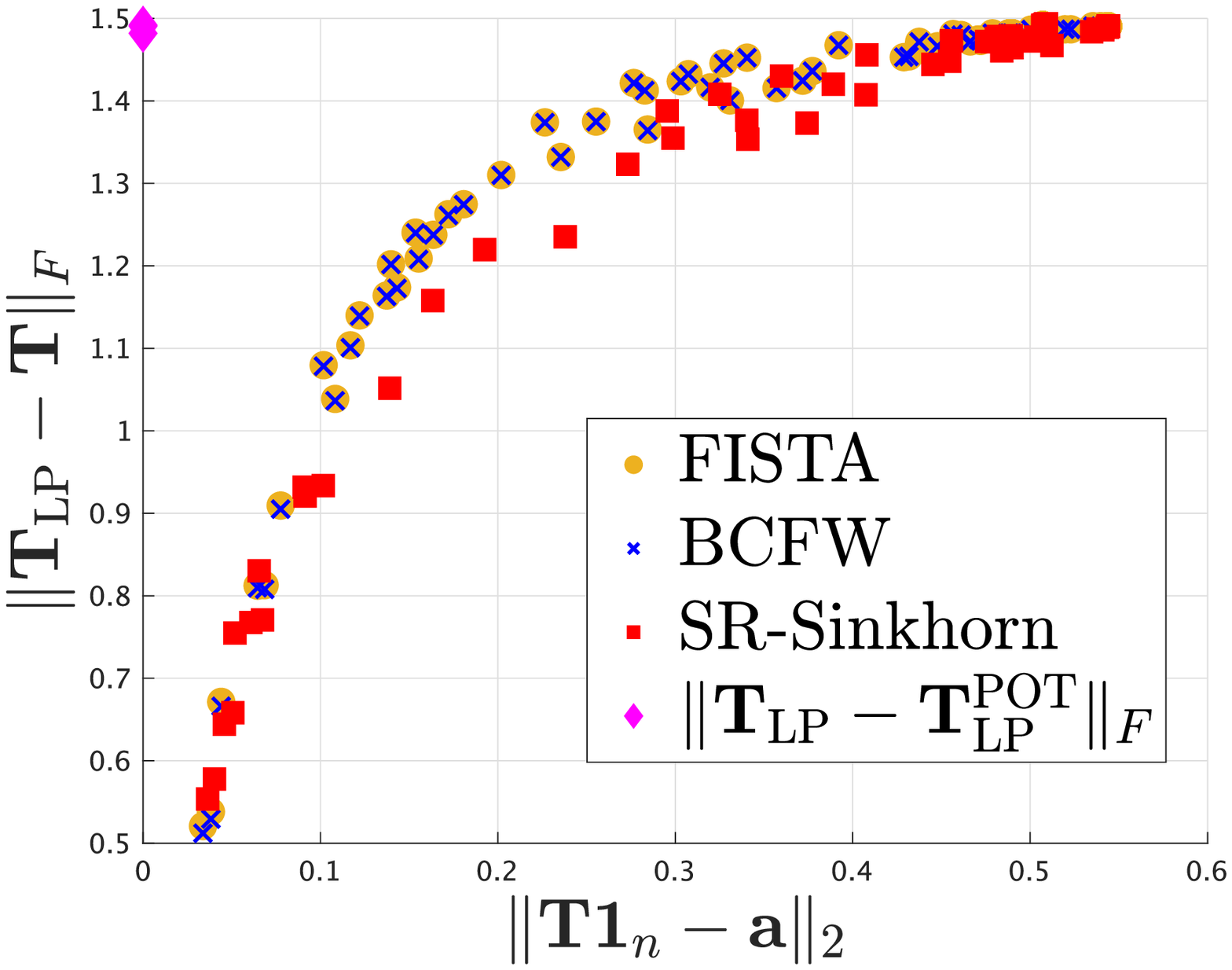}	
	
	{\small (b) deviation: $\| \mat{T}_{\rm LP}\!-\! \mat{T}\|_F$.}	
	\end{center}
	\end{minipage}	
\caption{Comparison with SROT algorithms.}	
\label{fig:CompSROTSolvers}	
\end{center}		
\end{figure}

We evaluate the empirical convergence performances of the SR--Sinkhorn algorithm by comparison with state-of-the-art algorithms including the fast iterative shrinkage-thresholding algorithm (FISTA) \cite{Beck_2009_SIAMIS} and the fast block--coordinate Frank--Wolfe algorithm (BCFW)\footnote{https://github.com/hiroyuki-kasai/SROT}  \cite{Fukunaga_ICASSP_2022,fukunaga2021fast} for the SROT problem. It must be emphasized that, in addition to the entropic regularization term, our formulation differs from those of these two algorithms because we use the KL divergence for $\Phi(\mat{T}\vec{1}_n,\vec{a})$ in (\ref{eq:SmoothSemiRelaxedOptimalTransport}) whereas the two others use $\frac{1}{2\lambda}\|\mat{T}\vec{1}_n-\vec{a}\|_2^2$. This difference prevents fair comparison under the same regularization parameters. Therefore, we measure two values: The distance $\langle \mat{T},\mat{C} \rangle$ and the transport matrix deviation $\| \mat{T}_{\rm LP} - \mat{T}\|_F$ at the end of the maximum iteration. The latter is deviation from the solution $\mat{T}_{\rm LP}$ of (\ref{eq:FormulationOptimalTransport}) obtained using the LP solver\footnote{\url{https://www.mosek.com/}.}. Subsequently, we plot these two values according to the marginal constraint gap $\|\mat{T}\vec{1}_n-\vec{a}\|_2$ under different $\tau$s and $\lambda$s. 
After randomly generating three synthetic datasets similarly to the mode described above, we set $\tau = 1/\lambda = (100,80, 60,40,20,10,8,6, \ldots, 0.0002, 0.0001)$. Also, $\eta$ is set $0.05$. The maximum iteration number is $1000$.
Figure \ref{fig:CompSROTSolvers}(a) presents the distance $\langle \mat{T},\mat{C} \rangle$, where $\langle \mat{T}_{\rm LP},\mat{C} \rangle$ is shown together at the vertical axis. Also, we plot $\langle \mat{T}_{\rm LP}^{\rm POT},\mat{C} \rangle$, where $\mat{T}_{\rm LP}^{\rm POT}$ is the solution of (\ref{Eq:Sinkhorn}) calculated using the LP solver. Figure \ref{fig:CompSROTSolvers}(b) portrays the transport matrix deviations. They all approach the LP solutions by strengthening the regularizer. They are getting closer to the POT problem when loosening the regularizer in an opposite way. From (b), it is apparent that the transport matrix from the SR--Sinkhorn algorithm algorithm is slightly closer to $\mat{T}_{\rm LP}$. Therefore, the KL-based regularization is apparently looser than the $\ell_2$-norm based ones. 

\end{document}